\PassOptionsToPackage{table}{xcolor}
\documentclass[11pt]{article}

\usepackage[preprint]{acl}

\usepackage{makecell}
\usepackage[T1]{fontenc}
\usepackage[utf8]{inputenc}
\usepackage{times}
\usepackage{latexsym}
\usepackage{inconsolata}
\usepackage{graphicx}
\usepackage{microtype}
\usepackage{hyperref}
\hypersetup{
  pdftitle={Ventriloquist LLMs: Linear Alignment of Late-Stage Representations},
  pdfauthor={Matt Gorbett, Suman Jana},
  pdfsubject={Machine Learning; Computation and Language},
  pdfkeywords={representation alignment, linear mapping, language models, representational convergence}
}
\usepackage{url}
\usepackage{booktabs}
\usepackage{amsmath,amssymb,amsthm}
\usepackage{amsfonts}
\usepackage{nicefrac}
\usepackage{xcolor}
\usepackage[most]{tcolorbox}
\usepackage{bm}
\usepackage{enumitem}
\usepackage{multirow}
\usepackage[acronym]{glossaries}
\usepackage{etoolbox}
\usepackage{titlesec}

\usepackage{CJKutf8}

\newacronym{dnn}{DNN}{deep neural network}
\newacronym{moe}{MoE}{mixture of experts}
\newacronym{lora}{LoRA}{low-rank adaptation}
\newacronym{rope}{RoPE}{rotary positional embeddings}
\newacronym{pca}{PCA}{principal component analysis}
\newacronym{cca}{CCA}{canonical correlation analysis}
\newacronym{he}{HE}{Homomorphic Encryption}
\newacronym{fhe}{FHE}{fully homomorphic encryption}
\newacronym{dp}{DP}{differential privacy}
\newacronym{sa}{SA}{secure aggregation}
\newacronym{rag}{RAG}{retrieval-augmented generation}
\newacronym{vdb}{VDB}{vector database}
\newacronym{lsh}{LSH}{locality sensitive hashing}
\newacronym{ddp}{DDP}{Distributed Data Parallel}
\newacronym{fsdp}{FSDP}{Fully Sharded Data Parallel}
\newacronym{nccl}{NCCL}{NVIDIA Collective Communications Library}
\newacronym{rpc}{RPC}{remote procedure call}
\newacronym{api}{API}{application programming interface}
\newacronym{eks}{EKS}{Elastic Kubernetes Service}
\newacronym{s3}{S3}{Amazon Simple Storage Service}
\newacronym{sql}{SQL}{structured query language}
\newacronym{ui}{UI}{user interface}
\newacronym{etl}{ETL}{extract, transform, load}
\newacronym{ftl}{FTL}{Federated Transfer Learning}
\newacronym{held}{HELIX}{Homomorphically Encrypted Linear Inference across models}
\newacronym{llm}{LLM}{Large language model}
\newacronym{lm}{LM}{language model}
\newacronym{ols}{OLS}{ordinary least-squares}
\newacronym{cka}{CKA}{Centered Kernel Alignment}

\newacronym{ood}{OOD}{out-of-distribution}
\newacronym{svcca}{SVCCA}{Singular Vector Canonical Correlation Analysis}

\newacronym{mpc}{MPC}{Secure Multi-Party Computation}
\makeglossaries

\theoremstyle{remark}

\setlist[itemize]{noitemsep, topsep=0pt}

\usepackage{titlesec}
\titlespacing*{\section}{0pt}{4pt}{2pt}
\titlespacing*{\subsection}{0pt}{3pt}{1pt}
\titlespacing*{\paragraph}{0pt}{2pt}{1em}

\usepackage[normalem]{ulem}

\usepackage{xcolor}
\definecolor{deltagreen}{HTML}{1B7F3A}
\definecolor{deltared}{HTML}{B82020}

\usepackage{times}
\usepackage{latexsym}

\usepackage[T1]{fontenc}

\usepackage[utf8]{inputenc}

\usepackage{microtype}

\usepackage{inconsolata}

\usepackage{graphicx}

\title{Ventriloquist LLMs: Linear Alignment of Late-Stage Representations}

\author{Matt Gorbett \\
  Capital One \\
  New York, NY 10003, USA \\
  \texttt{matthewgorbett@gmail.com} \\\And
  Suman Jana \\
  Department of Computer Science \\
  Columbia University \\
  New York, NY 10027, USA \\
  \texttt{sj2536@columbia.edu} \\}

\PassOptionsToPackage{final}{microtype}
\begin{document}

\maketitle

\begin{abstract}

Independently trained language models often learn compatible late-stage representations, despite differences in training objectives, architectures, and data modalities. We ask how far this compatibility extends: can a simple affine map let one model's hidden states be read directly by another model's output head? In this work, we learn affine transformations between the final hidden states of independent models and evaluate them across embedding classification, out-of-distribution detection, and autoregressive text generation. Across model pairs, we find that downstream performance is largely preserved under alignment, with linearly mapped source representations retaining both the decision boundaries and the confidence structure of a target model's classifier. Additionally, we show for the first time that linear alignment sometimes enables text generation across independently trained models, decoding one model's hidden states through another's frozen output head without fine-tuning. This capability is far from universal, and characterizing when it holds is a contribution of our work. We find that success is governed by two factors: tokenizer overlap, which correlates strongly with generation quality, and source-model scale, below which quality degrades sharply. Transfer is also asymmetric, with strong-to-weak mappings substantially outperforming weak-to-strong ones, indicating that source representational capacity, rather than the target head, is the limiting factor. Finally, we characterize where in the network this compatibility arises. A layer-wise analysis localizes this compatibility to the final one or two layers, indicating a late-stage, output-space phenomenon rather than evidence of shared reasoning.

\end{abstract}

\begin{figure}[t]
  \captionsetup{skip=0pt}
  \centering
  \includegraphics[width=\linewidth]{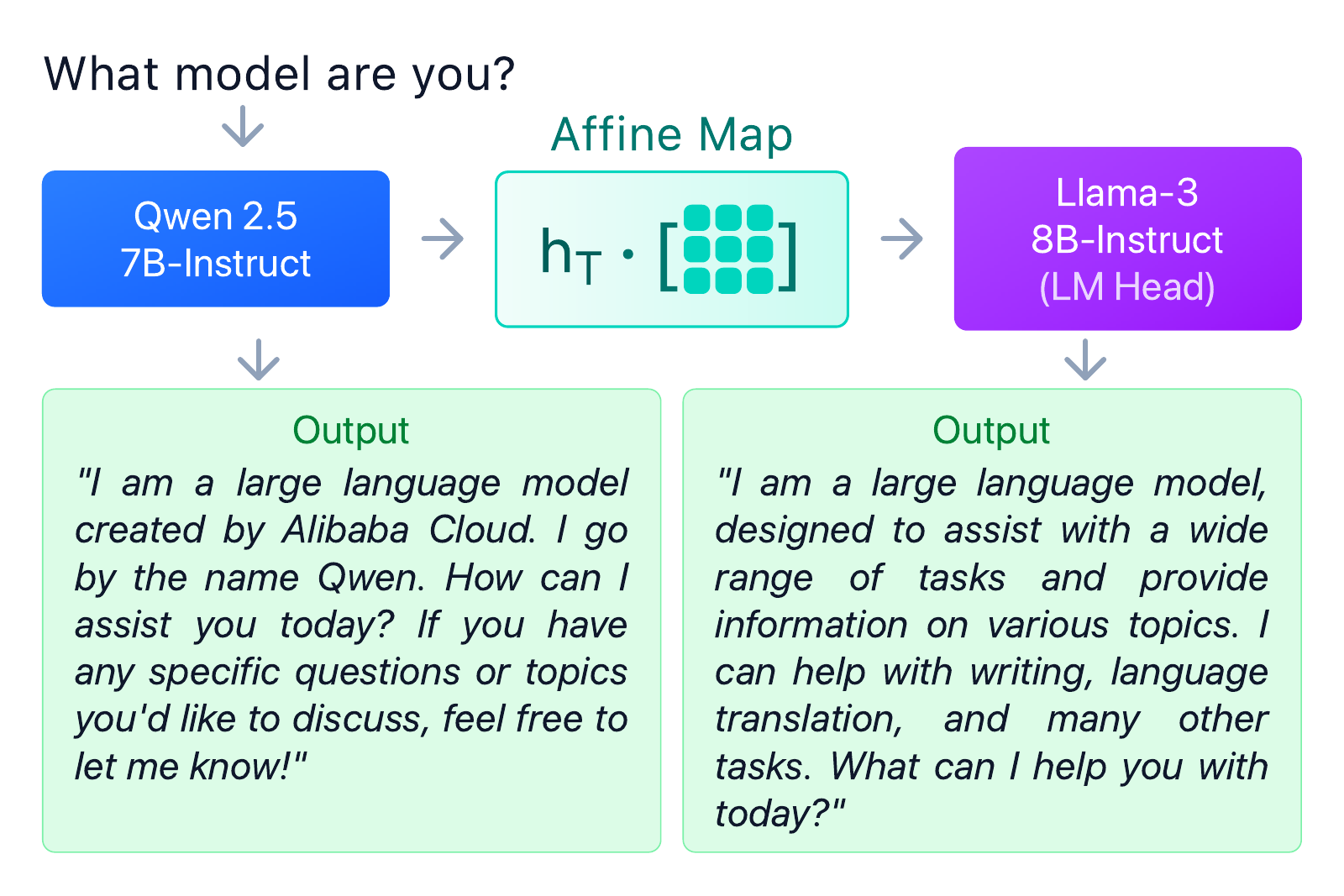}
  \caption{\textbf{Text Generation via Cross-Model Linear Alignment:} We learn an affine map from Qwen's final hidden states into Llama's feature space, enabling Qwen representations to be decoded by Llama's token head. The resulting hybrid model combines Qwen's embedding layer and transformer blocks with Llama's output head, producing coherent responses without adopting either model's identity.}
  \label{method_overview}
\end{figure}

\section{Introduction}
\glspl{llm} have become the standard paradigm for language understanding and generation, with both encoder-style and autoregressive architectures achieving strong generalization across diverse tasks \cite{srivastava2022beyond}. Their rapid progress is driven by scaling laws that link model size, compute, and data volume to emergent capabilities \cite{kaplan2020scaling, hoffmann2022training}. As \glspl{llm} continue to scale, recent evidence suggests that different models are also becoming more aligned in their learned representations \cite{huh2024platonic}, raising the possibility that independently trained systems share compatible output representations.

This emerging representational similarity enables new forms of model interoperability. For example, prior work on model stitching shows that independently trained models can be aligned through learned linear transformations, allowing downstream feature transfer across architectures \cite{BansalNB21}. Such compatibility supports multi-model inference pipelines, modular system design, and collaboration across heterogeneous systems \cite{jiang2024stitchinggptbert, chen2025transferring}. Yet it remains unclear how far this compatibility actually extends, which pairs of models it holds for, and where in the network it arises. Understanding when and where linear alignment succeeds is therefore both of practical interest for such pipelines and of scientific interest for what it reveals about the structure of independently learned representations.

In this work, we investigate the extent to which representational convergence in \glspl{llm} enables effective downstream inference via simple linear transformations across independently trained models. The key insight is that when models learn similar representations, their feature spaces can be aligned through a simple linear map (Figure~\ref{method_overview}). Specifically, we learn affine maps between the final hidden states of models with different architectures and training procedures, testing whether linear alignment is sufficient to preserve task performance on downstream linear heads when transferring from one model to another.

\begin{figure*} %
 \captionsetup{skip=0pt}
  \centering
  
  \includegraphics[width=1\linewidth]{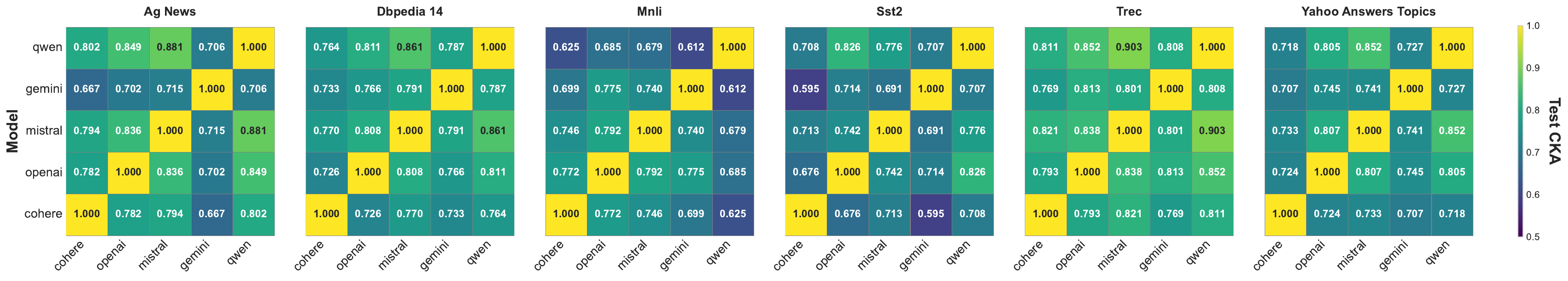} %

\caption{\textbf{Linear CKA similarity across embedding APIs.} We compute linear \gls{cka} \cite{kornblith2019similarity} on vendor-provided embeddings over shared inputs from multiple datasets. CKA values range from 0.595 to 0.881, indicating substantial shared linear structure across independently trained models.}
  \label{fig:heatmaps1}

\end{figure*}

To examine this question, we evaluate linear alignment across a diverse set of encoder-based and autoregressive \glspl{llm}. We first verify that independently trained models exhibit nontrivial shared linear structure (Figure~\ref{fig:heatmaps1}). We next measure alignment performance on supervised classification and \gls{ood} detection, observing minimal degradation across model pairs when using a fixed target linear head. Finally, we evaluate linear alignment on instruction-tuned generative models, showing that linear stitching can produce quality outputs (Figure~\ref{method_overview}): by mapping one model's penultimate hidden states into another's representation space and decoding with that second model's frozen output head, we obtain coherent generations without any fine-tuning. On the Alpaca dataset, we assess cross-model text generation using embedding similarity to native models, LLM-as-a-judge scoring, and human evaluation, finding that higher-similarity stitched pairs consistently yield higher-quality generations (Figure~\ref{fig:mmlu_similarity}, Table \ref{prompts}).

For text generation, we evaluate linear alignment across 34 model pairs using instruction-tuned generative models. Our investigation reveals two patterns: First, tokenizer compatibility strongly predicts success, with exact token match rate ($r = 0.898$) and Jaccard index ($r = 0.822$) correlating with text generation quality. Second, model scale imposes a minimum threshold, as all pairs with source models below 4B parameters produce lower quality results regardless of tokenizer alignment. We further observe that transfer is asymmetric: mapping from stronger to weaker models substantially outperforms the reverse, indicating that source representational capacity, rather than the target head, is the limiting factor. We assess quality through LLM-as-a-Judge evaluation, embedding similarity to source models, and human judgment.

Finally, we characterize \emph{where} in the network this compatibility arises. A layer-wise analysis shows that linear alignment fails almost entirely at early and middle layers and emerges only at the final one to two layers. This localization indicates that cross-model linear compatibility is a late-stage, output-space phenomenon rather than evidence of shared internal reasoning, and it clarifies the scope of our generation results: what transfers is the geometry of the output space, not deeper computation.

\begin{figure}[h!]
  \centering
  \includegraphics[width=0.9\linewidth]{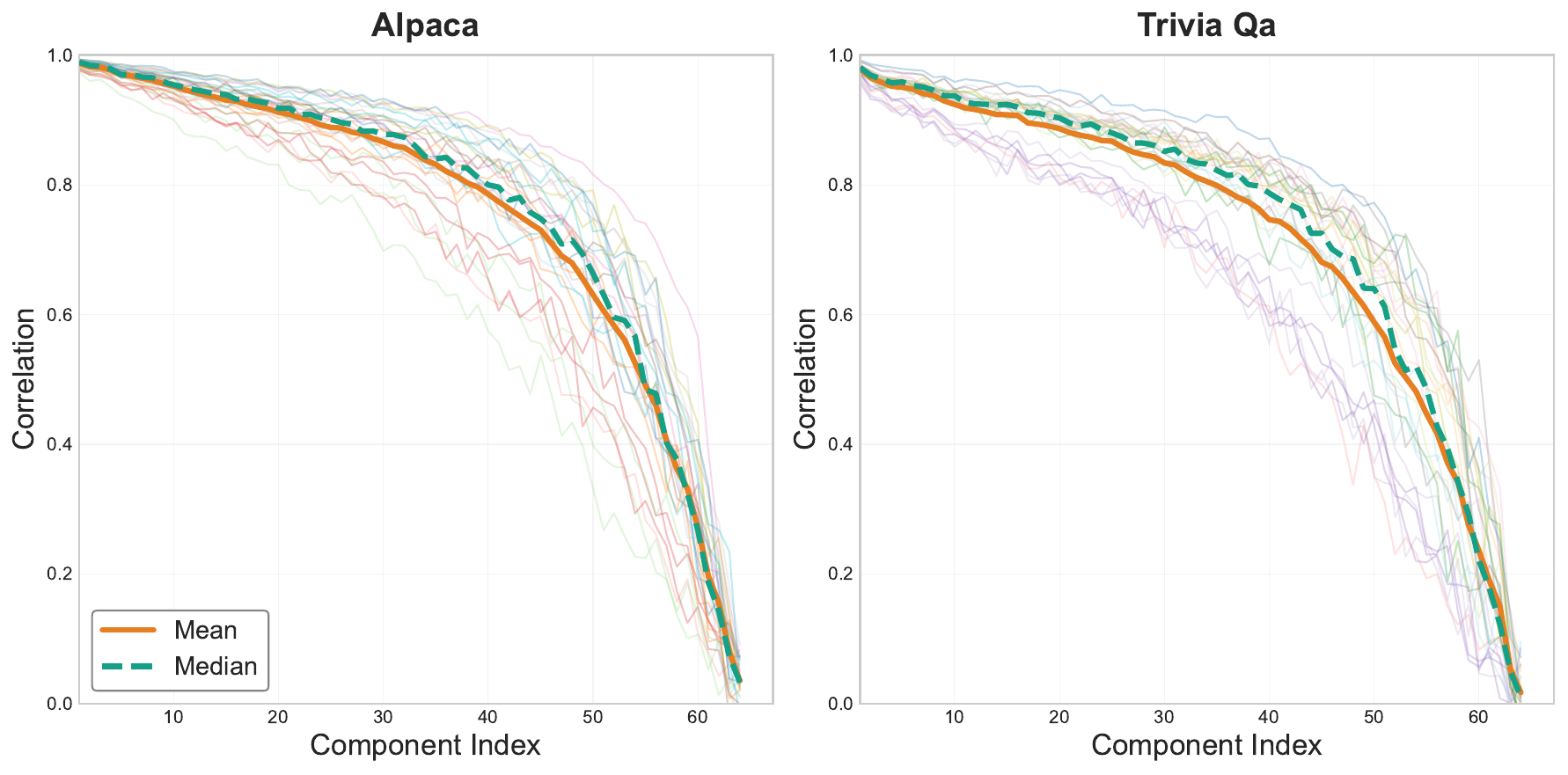} %
  \caption{SVCCA of eight instruction tuned \gls{llm} combinations at 64 components. Mean and median bolded.  }
  \label{fig:svcca}
\end{figure}

In summary, our contributions are as follows:

\begin{itemize}[
  nosep,
  topsep=2pt,
  partopsep=0pt,
  parsep=0pt,
  itemsep=0pt
]
    \item We characterize the extent to which a linear map can transfer knowledge between independently trained models on classification and \gls{ood} detection, and show for the first time that linear alignment in some cases supports \emph{text generation} across distinct \glspl{llm}. We identify tokenizer compatibility and source-model scale as the primary determinants of success, and observe an asymmetry in which strong-to-weak transfer substantially outperforms weak-to-strong.

    \item We localize where this compatibility lives via a layer-wise analysis, showing that linear alignment emerges only at the final one to two layers. This indicates that cross-model linear compatibility is a late-stage, output-space phenomenon rather than evidence of shared internal reasoning.

\end{itemize}

\section{Related Work} \label{background}

Understanding whether neural networks converge to similar representations despite stochastic training and non-convex optimization is a central question in machine learning \cite{li2015convergent, raghu2017svcca, klabunde2025similarity}.
A growing body of work shows that independently trained models often discover surprisingly aligned representations, particularly in overparameterized, high-capacity regimes.

\textbf{Representational Similarity.} \citet{kornblith2019similarity} introduced \gls{cka}, showing that identically structured CNNs trained from different seeds learn similar intermediate features. \citet{morcos2018insights} found that wider networks, and those that generalize better, exhibit stronger representational alignment.  At a higher level, the Platonic Representation Hypothesis \cite{huh2024platonic} proposes that large models increasingly converge toward a shared statistical understanding of the world, with similar geometric structure emerging across modalities and architectures.

\textbf{Model Stitching.} Building on representational convergence, model stitching asks whether a lightweight adapter can map intermediate representations from one model into another’s feature space while preserving downstream performance. Early CNN results show that shallow linear layers can stitch models trained under different settings \cite{lenc2015understanding, BansalNB21}. \citet{BansalNB21} argue that stitching complements statistical similarity metrics (e.g., CKA) by testing \textit{functional} interchangeability rather than mere geometric resemblance. Related work further shows that independently trained CNN and face-recognition models can be linearly aligned at the final layer with minimal accuracy loss \cite{mcneely2020inception, mcneely2022canonical}. Recent work extends stitching to transformers and \glspl{llm}: \citet{chen2025transferring} align hidden states across language models of different sizes via linear maps to transfer features. Unlike our approach, theirs focuses on computational efficiency. \citet{jiang2024stitchinggptbert} stitch autoregressive and bidirectional transformers (GPT and BERT) for look-ahead text understanding, supporting the feasibility of cross-\gls{llm} alignment.

\textbf{Linear Identifiability.} \citet{roeder2020linear} show that for a broad class of models, including supervised, contrastive, and causal language models, representations learned on the same data and architecture are \emph{linearly identifiable}: there exists an invertible matrix \( W \) such that \( Z_B \approx W Z_A \). This result provides a theoretical foundation for the use of linear alignment methods like ours. However, the identifiability theorem is guaranteed only when architectures, objectives, and data distributions match. When these conditions differ, representational equivalence may be approximate.

\label{related}

\section{Linear Alignment Characterization}
\label{analysis}

In this section, we test whether a simple linear map can align independently trained \glspl{llm} while preserving downstream task behavior.

\begin{table*}[t]
\centering
\small
\renewcommand{\arraystretch}{1}
\begin{tabular*}{\textwidth}{@{\extracolsep{\fill}}llccccc@{}}
\toprule
Target & Source & \multicolumn{2}{c}{Classification Acc.} &  & \multicolumn{2}{c}{OOD AUROC} \\
\cmidrule(lr){3-4} \cmidrule(lr){6-7}
 & & Baseline & Linear Map & OOD Dataset & Baseline & LinMap \\
\midrule
\rowcolor{gray!10}
\multicolumn{7}{c}{\textbf{SST-2 (50\%)}} \\
\midrule
Gemini  & OpenAI  & 94.5 & 93.1 & \multirow{4}{*}{AGNews} & 0.826 & 0.801 \\
Cohere  & Gemini  & 94.4 & 91.7 &                          & 0.819 & 0.870 \\
OpenAI  & Cohere  & 93.0 & 92.4 &                          & 0.875 & 0.843 \\
Mistral & Qwen    & 94.5 & 93.7 &                          & 0.859 & 0.826 \\
\midrule
\rowcolor{gray!10}
\multicolumn{7}{c}{\textbf{TREC (17\%)}} \\
\midrule
OpenAI  & Qwen    & 96.0 & 95.6 & \multirow{4}{*}{AGNews} & 0.738 & 0.916 \\
Cohere  & Gemini  & 97.0 & 91.2 &                          & 0.921 & 0.766 \\
OpenAI  & Cohere  & 96.4 & 94.4 &                          & 0.954 & 0.802 \\
Mistral & Qwen    & 97.0 & 96.6 &                          & 0.921 & 0.940 \\
\midrule
\rowcolor{gray!10}
\multicolumn{7}{c}{\textbf{AG News (25\%)}} \\
\midrule
Gemini  & OpenAI  & 92.6 & 91.6 & \multirow{4}{*}{MNLI}   & 0.908 & 0.891 \\
OpenAI  & Qwen    & 92.6 & 91.9 &                          & 0.953 & 0.913 \\
Cohere  & Gemini  & 91.9 & 91.5 &                          & 0.875 & 0.885 \\
Mistral & Qwen    & 92.6 & 92.4 &                          & 0.953 & 0.940 \\
\midrule

\rowcolor{gray!10}
\multicolumn{7}{c}{\textbf{DBpedia (7\%)}} \\
\midrule
Gemini  & OpenAI  & 98.9 & 98.6 & \multirow{5}{*}{Yahoo} & 0.971 & 0.975 \\
OpenAI  & Qwen    & 99.0 & 98.7 &                          & 0.979 & 0.974 \\
Cohere  & Gemini  & 98.9 & 98.7 &                          & 0.973 & 0.980 \\
OpenAI  & Cohere  & 98.9 & 98.7 &                          & 0.971 & 0.973 \\

\bottomrule
\end{tabular*}
\caption{\textbf{Downstream performance is preserved under task-supervised linear alignment.} We train a linear classifier $f_A(\cdot)$ on target embeddings $g_A(\cdot)$ (\textbf{Baseline}), then fit a linear map from source to target embedding space and apply $f_A$ to the mapped source embeddings (\textbf{Linear Map}). In-distribution classification accuracy and OOD detection AUROC are largely preserved across all model pairs.}
\label{tab:linmap}
\end{table*}

\paragraph{Preliminaries}
\label{sec:preliminaries}
For each experiment, we consider two embedding models: the target model $\mathcal{F}_A$ consists of a representation function $g_A:\mathcal{X}\rightarrow\mathbb{R}^{d_A}$ followed by a task head $f_A:\mathbb{R}^{d_A}\rightarrow\mathcal{Y}$. The source model $\mathcal{F}_B$ provides a representation function $g_B:\mathcal{X}\rightarrow\mathbb{R}^{d_B}$. Throughout, $g_A(\cdot)$ and $g_B(\cdot)$ correspond to the final hidden representations produced by their respective transformer encoders.

We focus on settings where the target head ($f_A$) is linear. For a $K$-class classification task, the head takes the form $f_A(z)=zV+c$, with parameters $V\in\mathbb{R}^{d_A\times K}$ and $c\in\mathbb{R}^{K}$ learned on labeled training data using $g_A(x)$. 

To relate the two representation spaces, we learn a linear alignment from the source to the target space. Given an input $x\in\mathcal{X}$, the aligned representation is $\hat{z}_A=g_B(x)W+b$, where $W\in\mathbb{R}^{d_B\times d_A}$ and $b\in\mathbb{R}^{d_A}$ are learned parameters. For all experiments, predictions are obtained by applying the fixed classifier $f_A(\cdot)$ to $\hat{z}_A$ without retraining.

In our experiments, $g_A$ and $g_B$ are instantiated using a set of pretrained language models drawn from both vendor-provided embedding services and locally hosted autoregressive models. For embedding models, we use OpenAI’s text-embedding-3-small, Cohere’s embedding API, Google’s Gemini embedding-001 model, Qwen3-Embedding-8B, and e5-mistral-7b-instruct models. In all cases, the representation functions $g_A(\cdot)$ and $g_B(\cdot)$ are kept fixed, and only the affine alignment parameters $(W,b)$ are learned.

\textbf{Representational Similarity.}
Before evaluating behavioral transfer, we first verify that independently trained models exhibit shared linear structure. Figure~\ref{fig:heatmaps1} reports representative CKA similarity across model pairs; full CKA/SVCCA methodology and results are deferred to the Appendix.

\subsection{Cross-Model Alignment to Downstream Tasks}
\label{sec:downstream_eval}

We next test whether a simple linear transformation between embedding models preserves downstream performance on supervised classification and \gls{ood} detection. Since embeddings are commonly used as features for linear classifiers \cite{devlin2018bert, tunstall2022setfit}, this provides a natural testbed for cross-model representation compatibility.

\textbf{Experimental setup.}
For each dataset, we designate a target model and a source model. We train the target linear classifier $f_A(\cdot)$ on the training split using target embeddings , and learn an affine map $(W^*,b^*)$ on the same split to project source embeddings $g_B(x)$ into the target feature space. At test time, we freeze $f_A(\cdot)$ and evaluate predictions using aligned source representations:
\[
\hat{y} \;=\; f_A\!\left(g_B(x)\,W^* + b^*\right).
\]

\paragraph{OOD detection.}
OOD detection evaluates whether a model can separate in-distribution inputs from unseen data by probing its logits \emph{confidence}. We use the \emph{Energy score} \cite{liu2021energy} for logits $f_A(z)\in\mathbb{R}^K$:

\begin{equation}
  \mathcal{E}(z) \;=\; -\log \sum_{k=1}^{K} \exp\bigl(f_A(z)_k\bigr)  
\end{equation}

Lower Energy indicates higher confidence, while higher values are characteristic of OOD inputs. We report AUROC by thresholding $\mathcal{E}(z)$ to distinguish in- vs.\ out-of-distribution samples. Since Energy depends on the full logit distribution, it provides a sensitive test of whether linear alignment preserves the target model’s confidence structure.

\paragraph{Results.}
We report in-distribution classification accuracy and OOD AUROC, where OOD samples come from an alternative dataset. \textbf{Baseline} trains and evaluates a linear classifier $f_A(\cdot)$ on target embeddings $g_A(\cdot)$, while \textbf{Lin.Map} applies the same classifier to linearly aligned source embeddings. Table~\ref{tab:linmap} shows that linear alignment largely preserves classification accuracy and achieves competitive OOD detection performance, with AUROC often matching or exceeding the baseline, indicating that the mapping recovers both decision boundaries and confidence structure.

\begin{table}[t]
\centering
\small
\setlength{\tabcolsep}{3pt}
\renewcommand{\arraystretch}{1.1}
\begin{tabular*}{\columnwidth}{@{\extracolsep{\fill}}llcccc@{}}
\toprule
\textbf{Model 1} & \textbf{Model 2} &
\multicolumn{2}{c}{\textbf{Base}} & \multicolumn{2}{c}{\textbf{Mapped}} \\
\cmidrule(lr){3-4}\cmidrule(lr){5-6}
& & M1 & M2 & 1$\rightarrow$2 & 2$\rightarrow$1 \\
\midrule
Llama3-8B    & Qwen2.5-7B  & 58 & 70 & 48 & 68 \\
Gemma3-270M  & Llama3-8B   & 22 & 58 & 20 & 49 \\
Gemma3-270M  & Qwen2.5-7B  & 22 & 70 & 21 & 68 \\
Llama3.2-1B  & Llama3-8B   & 42 & 58 & 36 & 58 \\
Llama3.2-1B  & Qwen2.5-7B  & 42 & 70 & 28 & 69 \\
\bottomrule
\end{tabular*}
\caption{\textbf{Cross-model linear alignment on MMLU.} Baseline shows native model accuracy; mapped shows accuracy after linearly transforming Model 1's representations to Model 2's head (M1$\rightarrow$M2), or vice versa. Mapping from stronger to weaker models preserves performance, while mapping from weaker to stronger degrades substantially.}
\label{tab:mmlu_alignment}
\end{table}

\subsection{Text Generation} 
We next evaluate whether linear alignment extends to the more demanding setting of autoregressive text generation. Unlike classification, which requires a single forward pass, generation demands coherent sequential predictions over many decoding steps where errors can accumulate, making generation a stringent test of whether linear maps preserve the structure necessary for multi-step inference.

\textbf{Experimental Setup.}
We evaluate cross-model generation using open-source instruction-tuned models ranging from 270M to 32B parameters. For each model pair, we learn an affine transformation that maps source model B's penultimate layer representations to target model A's prediction head space. The transformation is learned via ordinary least squares with ridge regularization ($\lambda = 10^{-4}$) on 4,000 training examples from either MMLU or Alpaca. We extract hidden states using token-level alignment based on character offsets. For each token in model A at position $i$ with character end offset $e_A^i$, we match it to the model B token whose end offset equals or immediately follows: $j = \arg\min_k \{e_B^k \mid e_B^k \geq e_A^i\}$. This produces aligned pairs $(h_A^i, h_B^j)$ for training the affine map, accommodating tokenizer differences without requiring shared vocabularies.

Crucially, we use model A's prediction head without fine-tuning. At each decoding step, we extract model B's penultimate hidden state, apply the learned affine transformation to map into model A's representation space, and pass the result to model A's frozen LM head $f_A(\cdot)$ for next-token prediction. This tests whether linear mapping alone can enable zero-shot cross-model generation. We use greedy decoding for a maximum of 128 tokens.

\begin{figure}[t]
\centering
\includegraphics[width=\columnwidth]{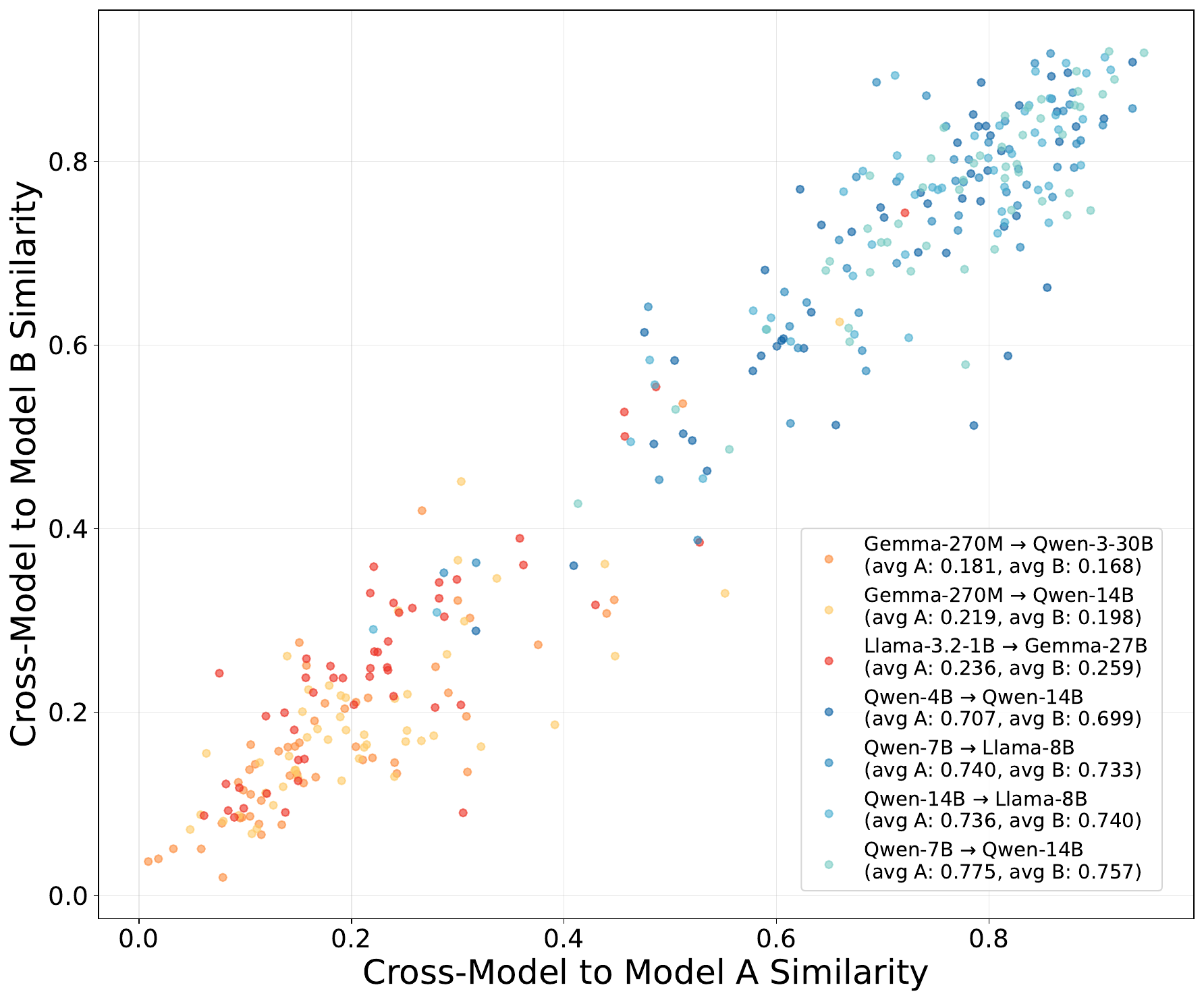}
\caption{\textbf{Cross-model embedding similarity to native models.} Each point represents an Alpaca test prompt. High-similarity pairs (upper right) produce coherent text, while low-similarity pairs (lower left) generate incoherent outputs.}
\label{fig:mmlu_similarity}
\end{figure}

\begin{figure*}[h!]
\centering
\captionsetup{skip=2pt}

\begin{tcolorbox}[
  enhanced,
  colback=blue!2,
  colframe=blue!60!black,
  boxrule=0.5pt,
  arc=2pt,
  left=4pt, right=4pt, top=2pt, bottom=2pt,
  title=\textbf{Olmo-7B $\rightarrow$ Qwen2.5-14B},
  fonttitle=\bfseries\small,
  coltitle=white,
  colbacktitle=blue!60!black,
  attach boxed title to top left={xshift=3pt, yshift=-1.5pt},
  boxed title style={
    colback=blue!60!black,
    sharp corners,
    boxrule=0pt,
    top=1pt, bottom=1pt, left=3pt, right=3pt,
  },
  before skip=3pt, after skip=2pt,
]
\small
\setlength{\parskip}{0pt} 

\textcolor{gray!80!black}{\itshape What are the benefits of social media for businesses?}\\
\textbf{\textcolor{blue!70!black}{Completion:}} The key benefits of using social media for businesses include increased brand awareness, improved customer engagement, enhanced advertising opportunities, better customer service, and cost-effective marketing\dots

\smallskip\hrule\smallskip

\textcolor{gray!80!black}{\itshape Evaluate the following expression: $(6+2)\times(2-2)$.}\\
\textbf{\textcolor{blue!70!black}{Completion:}} The result of the given expression is 0. First we calculate the addition and the subtraction, which are 8 and 0, and then the 8 is multiplied by 0\dots

\smallskip\hrule\smallskip

\textcolor{gray!80!black}{\itshape Describe how the given product works in 1--2 sentences: A portable charger.}\\
\textbf{\textcolor{blue!70!black}{Completion:}} A portable charger is a small, portable device with a high-capacity battery that recharges phones, tablets, or other electronics on the go. It has a set number of charge cycles and can recharge devices that\dots

\smallskip\hrule\smallskip

\textcolor{gray!80!black}{\itshape Summarize this paper: "Exploring Voice-Based Human-Robot Interaction in Non-Technical Domains"\dots}\\
\textbf{\textcolor{blue!70!black}{Completion:}} The research developed a conversational AI system enabling users to interact with a home-based robot using natural language, allowing intuitive assistance with household tasks in a non-technical, everyday setting.

\end{tcolorbox}

\vspace{2pt}

\begin{tcolorbox}[
  enhanced,
  colback=green!2,
  colframe=green!40!black,
  boxrule=0.5pt,
  arc=2pt,
  left=4pt, right=4pt, top=2pt, bottom=2pt,
  title=\textbf{Llama3-8B $\rightarrow$ Qwen2.5-14B},
  fonttitle=\bfseries\small,
  coltitle=white,
  colbacktitle=green!40!black,
  attach boxed title to top left={xshift=3pt, yshift=-1.5pt},
  boxed title style={
    colback=green!40!black,
    sharp corners,
    boxrule=0pt,
    top=1pt, bottom=1pt, left=3pt, right=3pt,
  },
  before skip=3pt, after skip=2pt,
]
\small
\setlength{\parskip}{1pt}

\textcolor{gray!80!black}{\itshape Provide advice to a person who is considering becoming an entrepreneur.}\\
\textbf{\textcolor{green!40!black}{Completion:}} First, make sure you have a solid business idea. This is the foundation of your business, and it's crucial that you have a clear and well-defined idea of what you want to do. Do your research, identify your target market, and make sure you have a unique value proposition that sets your business apart from\dots

\smallskip\hrule\smallskip

\textcolor{gray!80!black}{\itshape Write a short article about a new technology that you think will revolutionize the world.}\\
\textbf{\textcolor{green!40!black}{Completion:}} One new technology that has the potential to revolutionize the world is the development of a fully autonomous transportation system. This technology has been in the making for several years and is now on the verge of being fully implemented. The concept of a fully autonomous transportation system is one where vehicles are able to navigate and operate without the need for human input. This could be in the form of autonomous cars, drones\dots
\smallskip\hrule\smallskip

\textcolor{gray!80!black}{\itshape Research the differences between an object-oriented and a procedural programming language.}\\
\textbf{\textcolor{green!40!black}{Completion:}} In a procedural language, the program is organized as a series of procedures or functions that perform specific tasks, executed in a specific sequence; the focus is on the process. In contrast, an object-oriented language is based on the concept of objects that bundle data with behavior\dots

\end{tcolorbox}

\caption{Chat prompts and their corresponding completions for a given model pair using cross-model linear alignment.}
\label{prompts}
\end{figure*}

\subsubsection{MMLU: Structured Output Generation}

\textbf{Task and Evaluation.} We evaluate accuracy on 100 MMLU test questions, using greedy decoding with early stopping when a valid answer letter (A, B, C, or D) is generated. For comparison, we report baseline accuracy for each model's native performance on the same questions.

\textbf{Results.} Table~\ref{tab:mmlu_alignment} shows results for five model pairs. Mapping from stronger to weaker models preserves substantial accuracy, while mapping from weaker to stronger models degrades significantly. This asymmetric transfer demonstrates that strong representations retain semantic structure after linear transformation and can be decoded zero-shot by weaker models' prediction heads without fine-tuning. The asymmetry confirms that source representational capacity, not the target head, is a limiting factor.

\begin{figure*}[t]
  \centering
  \begin{minipage}{0.48\textwidth}
    \centering
    \includegraphics[width=\linewidth]{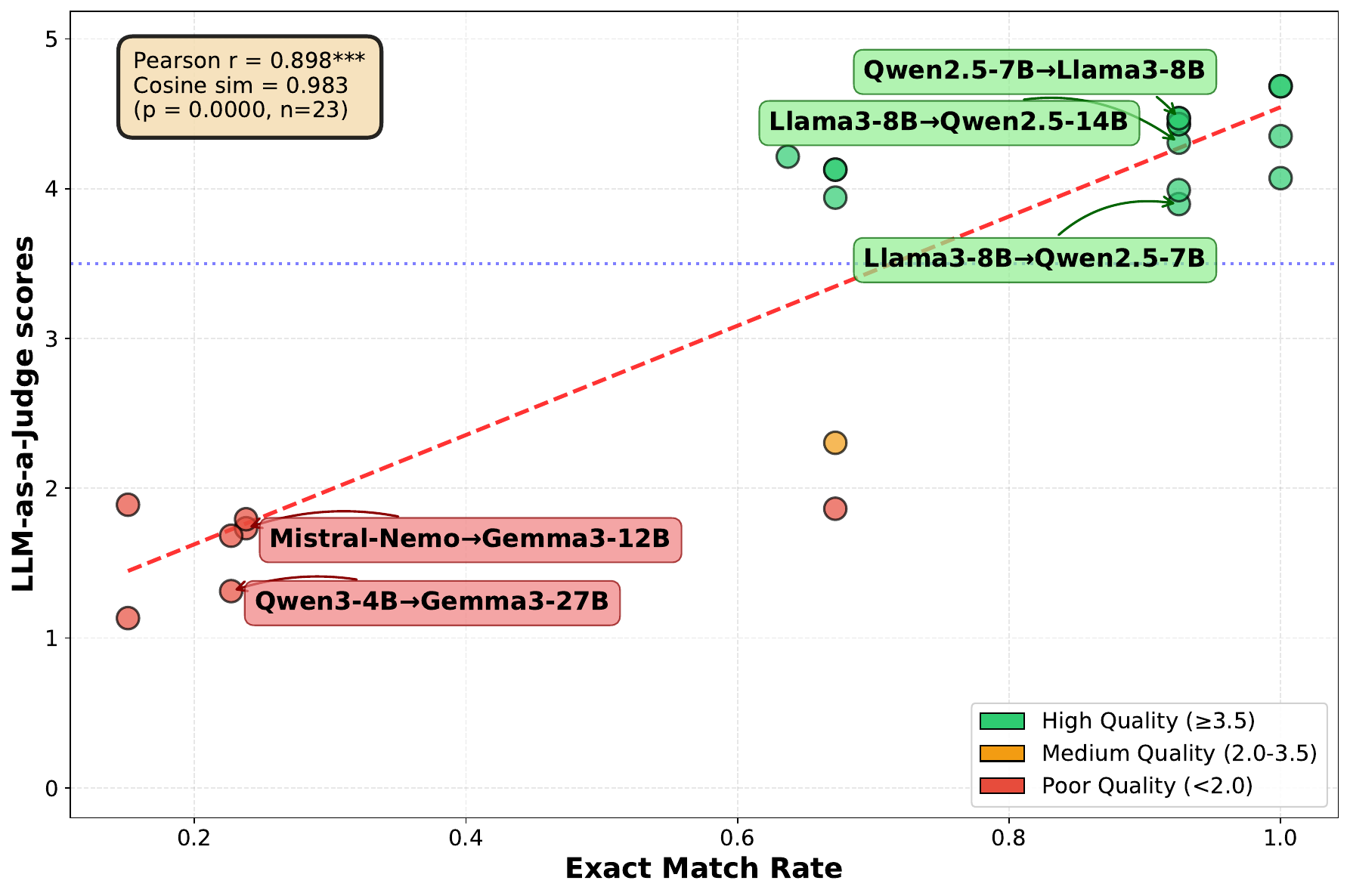}
  \end{minipage}
  \hfill
  \begin{minipage}{0.48\textwidth}
    \centering
    \includegraphics[width=\linewidth]{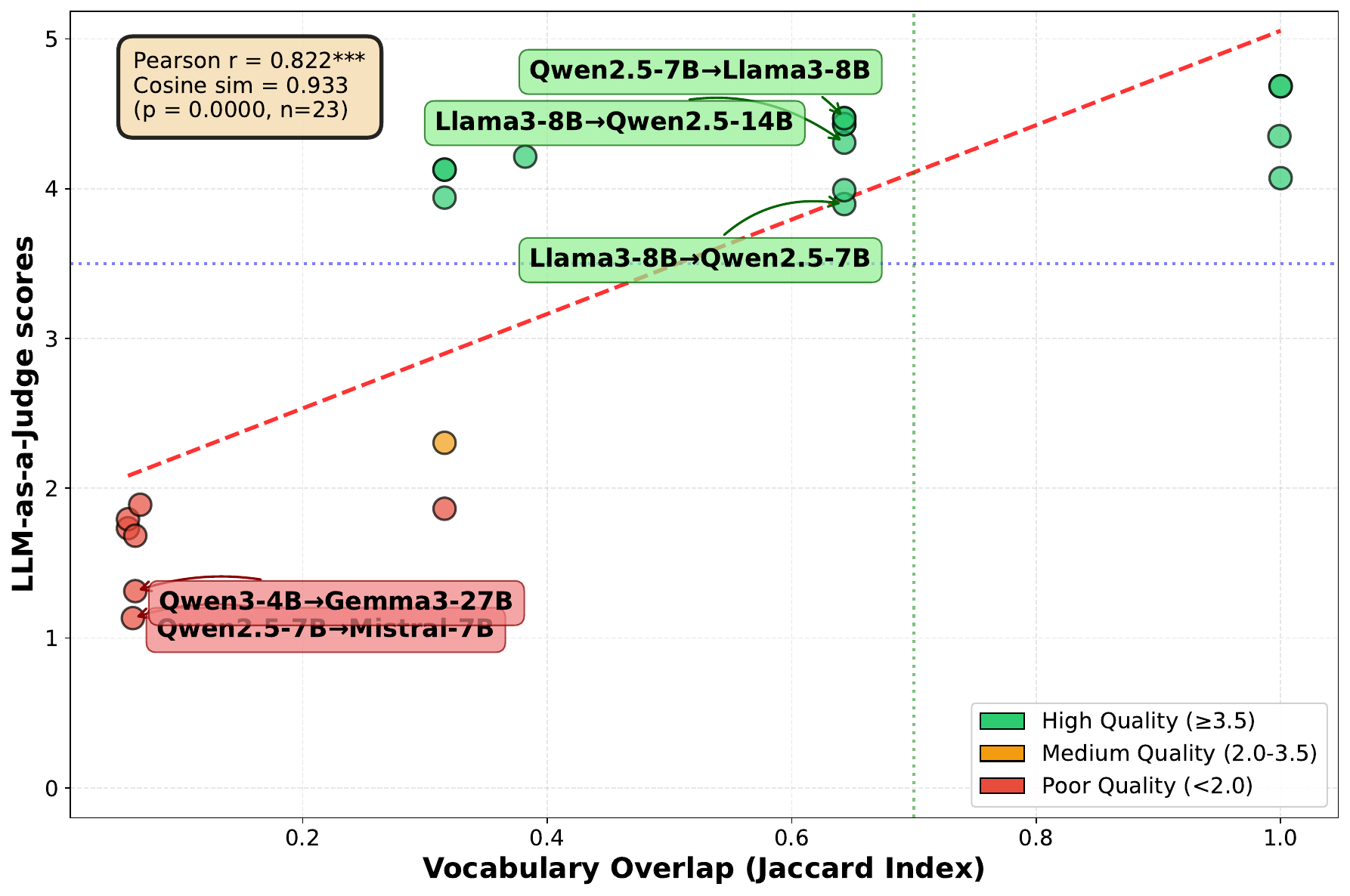}
  \end{minipage}
  \caption{\textbf{Tokenizer compatibility predicts cross-model generation quality.} \textit{Left:} Exact token match rate across 23 model pairs ($r=0.898$, $p<0.001$). \textit{Right:} Vocabulary overlap (Jaccard index) shows similar predictive power ($r=0.822$, $p<0.001$). }
  \label{fig:tokenizer_quality}
\end{figure*}

\subsubsection{Alpaca: Instruction Following}

\textbf{Task and Evaluation.} We evaluate cross-model generation on 100 held-out Alpaca test prompts using three complementary methods: (1) embedding similarity to native model outputs using OpenAI's text-embedding-3-small model; (2) LLM-as-a-Judge evaluation using GPT-5.1 and Claude-Sonnet-4, scoring relevance, quality, and helpfulness on a 1-10 scale; and (3) human evaluation on 200 samples (evenly split between single-model and cross-model) using a 1-5 scale based on proper English, quality, and coherence. %

\textbf{Results.} We find that certain model pairs align substantially better than others, with two consistent trends. First, smaller models ($\le$ 1B parameters) yield poor text generation performance when mapped to larger models, even when both models belong to the same family (Figure~\ref{fig:mmlu_similarity}). Second, \emph{tokenizer compatibility} is a predictor of cross-model generation success. Analyzing 23 larger pairs ($\geq$4B parameters), we find that exact token match rate (i.e. the fraction of tokens aligning at corresponding positions when tokenizing identical text) strongly correlates with generation quality ($r=0.898$, $p<0.001$; Figure~\ref{fig:tokenizer_quality}). High-quality pairs (LLM-judge score $\geq$3.5) consistently exhibit exact match $\geq$0.67, while failures show $\leq$0.24. Additionally, we analyze vocabulary overlap via Jaccard index, which shows similar predictive power ($r=0.822$, $p<0.001$). For example, Gemma's poor cross-family performance stems from low tokenizer compatibility (exact match $<$0.23, Jaccard $<$0.07). We find that pairs with exact match $>$0.7 succeed consistently, as evidenced by the clustering patterns in Figure~\ref{fig:mmlu_similarity}. Full results are in Appendix \ref{text_gen_exp}.

\textit{Embedding similarity.} Figure~\ref{fig:mmlu_similarity} reveals a strong correlation between embedding similarity and generation quality. Model pairs with consistent high quality text generation cluster into either high-similarity (upper right, cosine similarity $>0.7$ to both models), and model pairs with poor text generation cluster into low-similarity (lower left, $<0.3$) regions. High-similarity pairs like Qwen-7B$\leftrightarrow$Meta-8B and Qwen-14B$\rightarrow$Meta-8B produce outputs semantically similar to both native models, while low-similarity pairs like Gemma-270M$\rightarrow$Qwen-14B generate poor quality text. %

\textit{LLM-as-a-Judge scores.} Cross-model generation via linear alignment achieves LLM-judge scores of 4.0-4.7 for high-compatibility pairs (Qwen $\rightarrow$ Llama, Mistral-Nemo $\rightarrow$ Apertus), retaining 60-70\% of single-model baseline. Low-compatibility pairs such as Gemma and small models ($<$2B) produce incoherent text (scores $<$ 2.0), validating our embedding similarity analysis. Detailed scores and comparisons to single-model baselines are provided in the Appendix.

\textit{Human evaluation.} Human evaluation scores corroborate the LLM-as-a-Judge findings. Cross-model pairs judged as low-quality by the LLM received an average human score of 1.1 ($\pm$0.2), while high-quality cross-model pairs scored 3.0 ($\pm$0.7). For comparison, low-quality baseline models scored 2.5 ($\pm$0.8) and high-quality baseline models scored 4.5 ($\pm$0.3). Human evaluators ranked outputs in the same order as the automated judge, validating that embedding similarity and LLM-as-a-Judge scores as reliable quality metrics.

\textit{Perplexity Analysis.} Finally, we observe that larger model pairs (7--8B parameters) show lower perplexity degradation ($26.07\% \pm 12.26\%$) compared to smaller-to-larger mappings ($37.25\% \pm 25.84\%$), though small sample sizes ($N=3$--4) limit statistical conclusions.

\section{Where Does Linear Alignability Live?}
\label{sec:layer-depth}

The main experiments align post-final-norm hidden states. Is depth a free
choice, or is the cross-model linear correspondence concentrated in a
specific region of the residual stream? We answer with a depth sweep on
Qwen2.5-7B-Instruct ($L{=}28$, $d{=}3584$) $\leftrightarrow$
Meta-Llama-3-8B-Instruct ($L{=}32$, $d{=}4096$) in both directions.

\paragraph{Setup.} For each layer pair we extract per-token hidden states
on $4{,}000$ MMLU train questions ($\sim$$1.55\textsc{m}$ aligned token
pairs), solve OLS ridge for the map $W$, and evaluate by greedy mapped
generation through the target LM head on $100$ MMLU test questions. The
ridge residual reaches machine epsilon ($\sim$$10^{-12}$) in every
configuration, so accuracy reflects what is recoverable from the source
representations, not solver quality. We sweep six depths per direction:
three intermediate blocks ($\sim$$L/4, L/2, 3L/4$), the two pre-final-norm
blocks ($h[-2], h[-1]$), and the post-final-norm
\texttt{last\_hidden\_state} used in the main paper.

\paragraph{Results.} Table~\ref{tab:layer-depth} shows that linear
alignability is effectively a step function in depth. Intermediate layers
fit the training pairs to machine precision yet decode to $0\%$ accuracy,
emitting rare-vocab fragments rather than coherent text; accuracy stays flat
at zero through $3L/4$ and then jumps only at the final blocks. The
phenomenon is thus a property of the \emph{top} of the residual stream: no
shallower layer admits a cheap, unrelated linear map. The final RMSNorm is
absorbed by the mapper for free, as pre-norm $h[-1]$ and post-norm
\texttt{last\_hidden\_state} tie exactly in both directions, justifying the
main-paper choice. Pulling back one further block is direction-asymmetric,
a mild $70\!\to\!67$ wobble for Qwen$\to$Llama but a $58\!\to\!0$ collapse
for Llama$\to$Qwen, consistent with the expanding map being more forgiving
than the lossy reverse.

\begin{table}[!h]
  \centering
  \small
  \setlength{\tabcolsep}{4pt}

  \begin{tabular}{l cc cc}
    \toprule
    \multirow{2}{*}{Layer}
      & \multicolumn{2}{c}{Qwen $\to$ Llama}
      & \multicolumn{2}{c}{Llama $\to$ Qwen} \\
    \cmidrule(lr){2-3}\cmidrule(lr){4-5}
      & $(l_A,l_B)$ & Acc.\ & $(l_A,l_B)$ & Acc.\ \\
    \midrule
    $\sim L/4$          & (7, 8)   & 0.00 & (8, 7)   & 0.00 \\
    $\sim L/2$          & (14, 16) & 0.00 & (16, 14) & 0.00 \\
    $\sim 3L/4$         & (21, 24) & 0.00 & (24, 21) & 0.00 \\
    $h[-2]$             & (27, 31) & 0.67 & (31, 27) & 0.00 \\
    $h[-1]$             & (28, 32) & \textbf{0.70} & (32, 28) & \textbf{0.58} \\
    \texttt{last\_hidden} & --     & \textbf{0.70} & --       & \textbf{0.58} \\
    \bottomrule
  \end{tabular}
    \caption{Layer-depth ablation on Qwen2.5-7B-Instruct $\leftrightarrow$
  Meta-Llama-3-8B-Instruct. Linear ridge mapper, $4{,}000$ MMLU train
  questions ($\sim$$1.55\textsc{m}$ token pairs), evaluated on 100 MMLU
  test questions via greedy mapped generation through the target LM head.
  OLS residual at machine epsilon in every row.}
  \label{tab:layer-depth}
\end{table}

\section{Alignment Under Distribution-Mismatched Data} \label{experiments}

Section~\ref{analysis} fits the affine map $W^*$ on the evaluation task's own distribution. This leaves open what the map encodes: does alignment require task-relevant data, or is the source-to-target mapping a property of the two representation spaces recoverable from arbitrary text? If output-layer compatibility is model-intrinsic, a map fit on unrelated data should still transfer. We test this by fitting $W^*$ on out-of-distribution data and measuring retained classification accuracy. Classification is a clean testbed since the target classifier $f(\cdot)$ is fixed, so any accuracy change is attributable to the map alone \cite{devlin2018bert, wang2022text}. We use five embedding models (Google, OpenAI, Cohere, \textsc{E5-Mistral-7B}, \textsc{Qwen3-Embedding-8B}) across four datasets: TREC, MNLI, DBpedia, and AG News.

\begin{table}[t]
\centering
\small
\setlength{\tabcolsep}{4pt}
\renewcommand{\arraystretch}{1}
\begin{tabular*}{\columnwidth}{@{\extracolsep{\fill}}llcccccc@{}}
\toprule
& & \multicolumn{1}{c}{Full} & \multicolumn{2}{c}{Baseline} & \multicolumn{1}{c}{OOD} & \multicolumn{2}{c}{OOD + $k$-shot} \\
\cmidrule(lr){4-5} \cmidrule(lr){7-8}
Target & Source & Data & 64 & 128 & Only & 64 & 128 \\
\midrule
\multicolumn{8}{c}{\textit{TREC (17\%)}} \\
\midrule
Gemini  & OpenAI  & 95.4 & 56.6 & 77.6 & 57.6 & \underline{78.0} & \textbf{81.0} \\
OpenAI  & Qwen    & 96.4 & 59.0 & \underline{82.8} & 58.4 & 77.6 & \textbf{83.6} \\
Mistral & Qwen    & 96.6 & 59.4 & 82.2 & 65.6 & \underline{83.8} & \textbf{87.4} \\
\midrule
\multicolumn{8}{c}{\textit{MNLI (33\%)}} \\
\midrule
Gemini  & OpenAI  & 65.0 & 33.5 & 35.1 & 44.5 & \underline{48.5} & \textbf{48.9} \\
OpenAI  & Qwen    & 62.4 & 57.7 & 61.9 & 64.0 & \underline{64.9} & \textbf{65.2} \\
Mistral & Qwen    & 77.6 & 57.6 & 61.8 & 58.8 & \underline{71.3} & \textbf{72.9} \\
\midrule
\multicolumn{8}{c}{\textit{DBpedia (7\%)}} \\
\midrule
Gemini  & OpenAI  & 99.1 & 65.0 & 77.8 & 53.5 & \underline{89.3} & \textbf{94.1} \\
OpenAI  & Qwen    & 99.1 & 66.0 & 83.4 & 66.7 & \underline{91.0} & \textbf{95.8} \\
Mistral & Qwen    & 99.2 & 72.9 & 93.2 & 79.6 & \underline{93.6} & \textbf{96.5} \\
\midrule
\multicolumn{8}{c}{\textit{AG News (25\%)}} \\
\midrule
Gemini  & OpenAI  & 92.1 & 79.6 & 87.0 & 85.9 & \underline{88.4} & \textbf{88.7} \\
OpenAI  & Qwen    & 92.7 & 80.0 & 87.1 & 79.7 & \underline{88.6} & \textbf{89.2} \\
Mistral & Qwen    & 93.4 & 84.4 & 86.5 & 71.4 & \underline{87.9} & \textbf{88.2} \\
\bottomrule
\end{tabular*}
\caption{\textbf{Cross-model alignment classification accuracy (\%) under distribution-mismatched alignment data.} We fit the alignment map on a generic out-of-distribution corpus (\textbf{OOD Only}) and on that corpus plus $k$ in-distribution examples (\textbf{OOD + $k$-shot}), then classify with the fixed classifier $f(\cdot)$. \textbf{Full Data} is $f(\cdot)$ trained on the target's full labeled set; \textbf{Baseline 64/128} are classifiers trained on only the $k$ examples. The OOD-only map uses no task labels yet recovers most of the full-data accuracy. Best in \textbf{bold}, second-best \underline{underlined}; $\pm 1.5$ pp across three seeds.}
\label{crossmodel_results}
\end{table}

\textbf{Setup.} For each dataset, a target model's embeddings and labels train the classifier $f(\cdot)$ on the full training set, and a source model's representations are aligned into the target space. We estimate $W^*$ under three regimes: \textbf{OOD Only} uses a generic corpus (Wikipedia, IMDB) with no shared labels or task structure; \textbf{OOD + $k$-shot} adds $k\in\{64,128\}$ in-distribution examples. We report the target's own full-labeled-set accuracy (\textbf{Full Data}) and classifiers trained on only the $k$ examples (\textbf{Baseline $k$}) as reference points.

\textbf{Results.} Table~\ref{crossmodel_results} shows that alignment recovers strong accuracy even when $W^*$ is fit on out-of-distribution data. The OOD-only map, which sees no labeled task example, surpasses the 64-shot baseline in many configurations and on several pairs approaches the full-data ceiling (e.g., 79--87\% on AG News vs.\ 92--93\% full-data). Adding a few in-distribution examples closes most of the gap, with OOD + 128-shot matching or exceeding the corresponding few-shot baseline. That a purely out-of-distribution map transfers most of the way indicates the affine relationship between two models' output spaces is largely task-independent, consistent with it being a late-stage geometric property rather than something learned per task.

\section{Conclusion}

We characterize the extent to which independently trained \glspl{llm} admit linear alignment at their output layers, showing that affine maps between final hidden states preserve downstream performance on classification, out-of-distribution detection, and, in some cases, autoregressive text generation. Success is governed by tokenizer compatibility and source-model scale, transfer is asymmetric from stronger to weaker models, and a layer sweep localizes the effect to the final one or two layers, indicating that cross-model linear compatibility is a late-stage, output-space phenomenon rather than evidence of shared reasoning.

Future work includes developing tokenizer-agnostic methods to reach models with disjoint vocabularies, and exploring non-linear maps that may recover quality where linear ones fail.

\bibliography{custom}

\newpage
\appendix
\onecolumn
\section*{Appendix}

\section{\gls{llm} Similarity}
To assess the extent to which independently trained embedding models learn compatible linear structure, we analyze representational similarity using two complementary metrics: linear Centered Kernel Alignment (CKA) and Singular Vector Canonical Correlation Analysis (SVCCA). Both measures operate directly on representation matrices computed over a shared set of inputs and quantify the degree of shared linear geometry across models.

\subsection{CKA Heatmaps }

Our primary similarity analysis uses linear \gls{cka} \cite{kornblith2019similarity}, which measures similarity between two representation matrices
$Z_A \in \mathbb{R}^{n \times d_1}$ and
$Z_B \in \mathbb{R}^{n \times d_2}$
computed over the same set of $n$ inputs. Linear CKA is invariant to isotropic scaling and orthogonal transformations, making it a stable and widely used metric for comparing internal representations across models with different dimensionalities or parameterizations.

Formally, linear CKA is defined as
\vspace{-0.5em}
\[
\mathrm{CKA}(Z_A, Z_B)
= \frac{\|Z_A^\top Z_B\|_F^2}
       {\|Z_A^\top Z_A\|_F \, \|Z_B^\top Z_B\|_F },
\]
where $\|\cdot\|_F$ denotes the Frobenius norm. A CKA value of $1$ indicates identical representations up to orthogonal transformation and scaling, while a value of $0$ indicates no shared linear structure.

\paragraph{Representation extraction.}
We compute $Z_A$ and $Z_B$ using a mixture of deployed embedding APIs and locally run autoregressive \glspl{llm}. For embedding APIs, we use the provider’s default embedding outputs, which already reflect model-specific pooling and normalization strategies. For locally run autoregressive models, which are trained primarily for next-token prediction and often instruction fine-tuned, we adopt mean pooling over the final hidden layer to obtain fixed-dimensional sequence representations.

Concretely, given final-layer token embeddings $\{h_i\}_{i=1}^L$ for an input sequence of length $L$, we compute the sequence-level embedding as
\[
z = \frac{1}{L}\sum_{i=1}^{L} h_i.
\]
Prior work has shown that pooling hidden states of autoregressive models yields semantic representations competitive with those of dedicated embedding models \cite{zhang2024should, tang2024pooling}, making this a reasonable and widely adopted choice for cross-model comparison.

\textbf{Experimental Procedure.}
We compute linear CKA between pairs of embedding models using representations extracted on shared training and test splits. Embeddings are mean-centered prior to CKA computation. For embedding APIs, we use the provider’s default outputs, while for instruction-tuned autoregressive models we apply mean pooling over final-layer token embeddings.

CKA is computed on both splits following \cite{kornblith2019similarity}, using 5{,}000 training and 2{,}000 test samples. Each experiment is repeated three times with different random subsamples, and results are averaged. 

\begin{figure}[h!]
 \captionsetup{skip=0pt}
  \centering
  
  \includegraphics[width=1\linewidth]{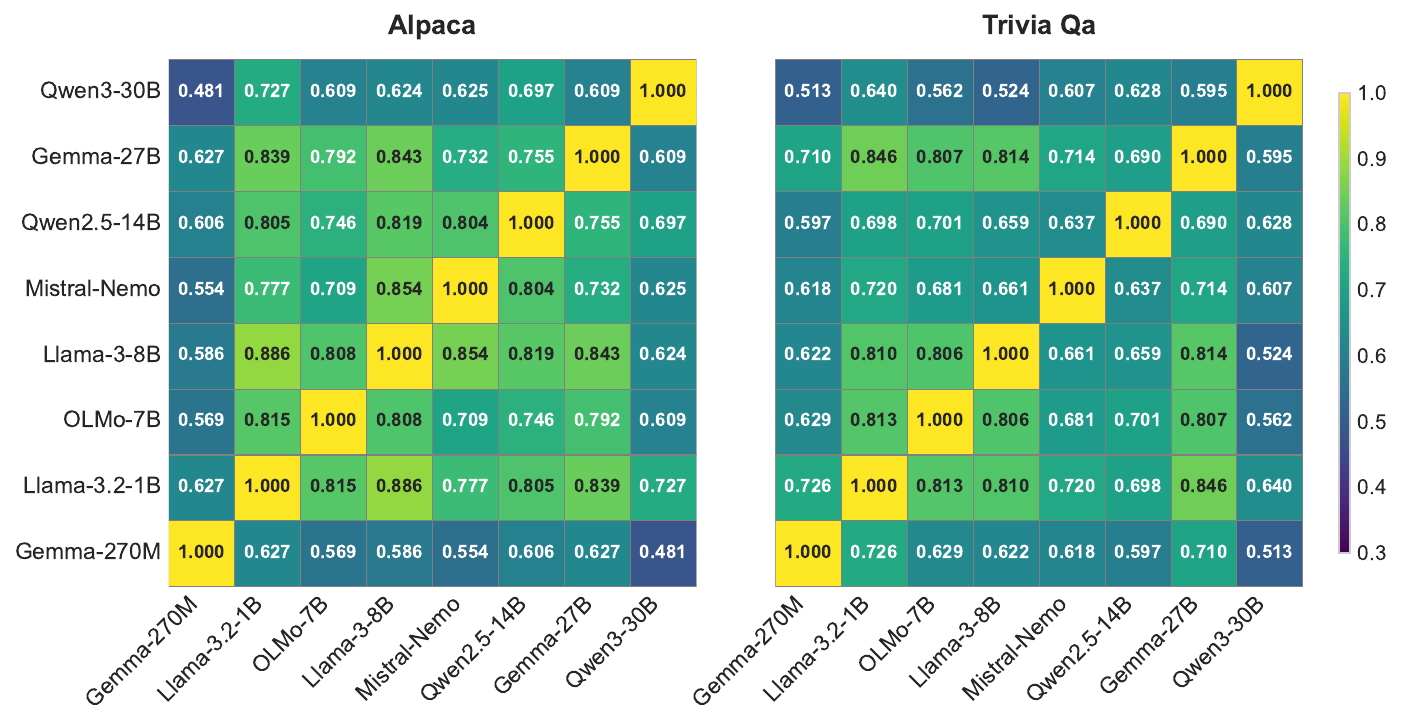} %
  \vspace{-8pt}
  \caption{Test-set \gls{cka} between embeddings across Alpaca and Trivia QA datasets.  Model embeddings were mean pooled at their final hidden state prior to calculating \gls{cka} with other models.  }
  \label{fig:heatmaps2}
\end{figure}

\paragraph{Results.}
Figure~\ref{fig:heatmaps2} reports linear \gls{cka} similarity across instruction-tuned autoregressive \glspl{llm}, computed on Alpaca and TriviaQA inputs. Each heatmap reports average test-set CKA computed using 5{,}000 samples to estimate representation covariance, with evaluation performed on 2{,}000 held-out samples. Results are averaged over three random subsampling runs to reduce variance.

We observe moderate to high CKA similarity across most autoregressive model pairs, indicating substantial shared linear structure despite differences in architecture, pretraining data, and instruction-tuning procedures. While variability across pairs is more pronounced than in encoder-style embedding models, many pairs still exhibit CKA values above $0.5$, which corresponds to meaningful shared linear structure and suggests strong potential for linear alignment. We hypothesize that the increased variance arises from task-specific shifts introduced during instruction tuning, which may alter representation geometry while preserving a common semantic backbone.

\subsection{SVCCA}
To complement CKA, we analyze representations using \gls{svcca} \cite{raghu2017svcca}, which measures correlation between low-rank subspaces extracted via singular value decomposition followed by canonical correlation analysis. Unlike CKA, which evaluates global similarity between full representation matrices, SVCCA emphasizes shared informative subspaces, making it particularly relevant for assessing the feasibility of linear alignment.

\textbf{SVCCA Experimental Procedure}
For each model pair, we extract embeddings on shared training and test splits. Encoder-style embedding models use provider-default outputs, while instruction-tuned autoregressive models use mean pooling over final-layer token embeddings.

SVCCA is fit on the training embeddings by first applying PCA to each model’s representations, followed by canonical correlation analysis (CCA) on the reduced features. We evaluate SVCCA on both the training and test splits using the fitted PCA and CCA transforms. We report results using fixed PCA dimensionalities of 64 and 128 components.

To ensure comparability, embeddings are aligned by truncating to the minimum number of available samples across models, with 10{,}000 training examples and 2{,}000 test examples. Each experiment is repeated three times with different random subsamples, and reported correlations are averaged across runs. As a control, we compute a random baseline by shuffling one model’s embeddings prior to SVCCA.

\begin{figure}[h] %
  \centering
  \includegraphics[width=.95\linewidth]{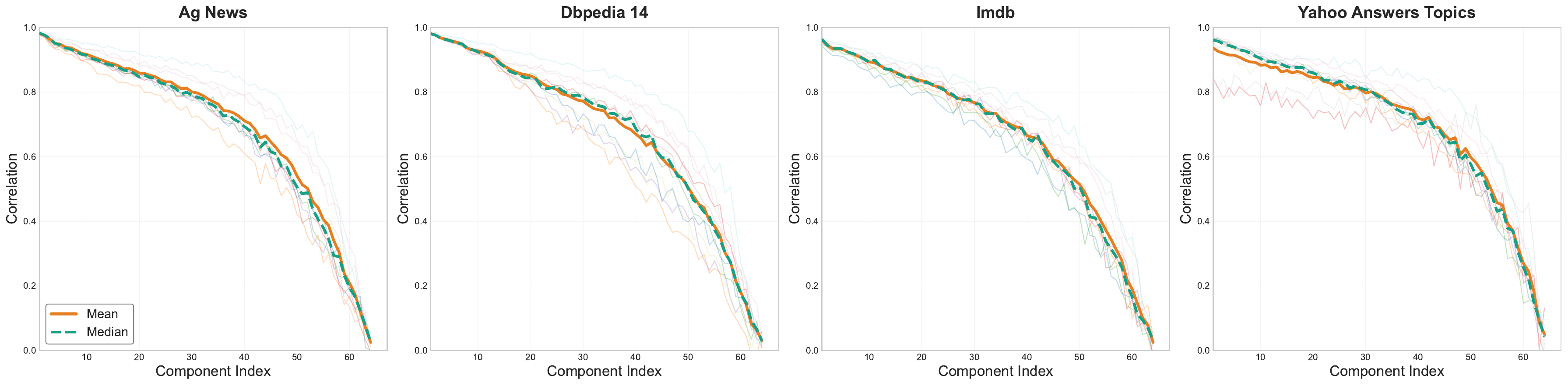} %
  \caption{SVCCA of four datasets on five embedding model combinations. }
  \label{fig:svcca1}
\end{figure}

\begin{figure}[h] %
  \centering
  \includegraphics[width=.95\linewidth]{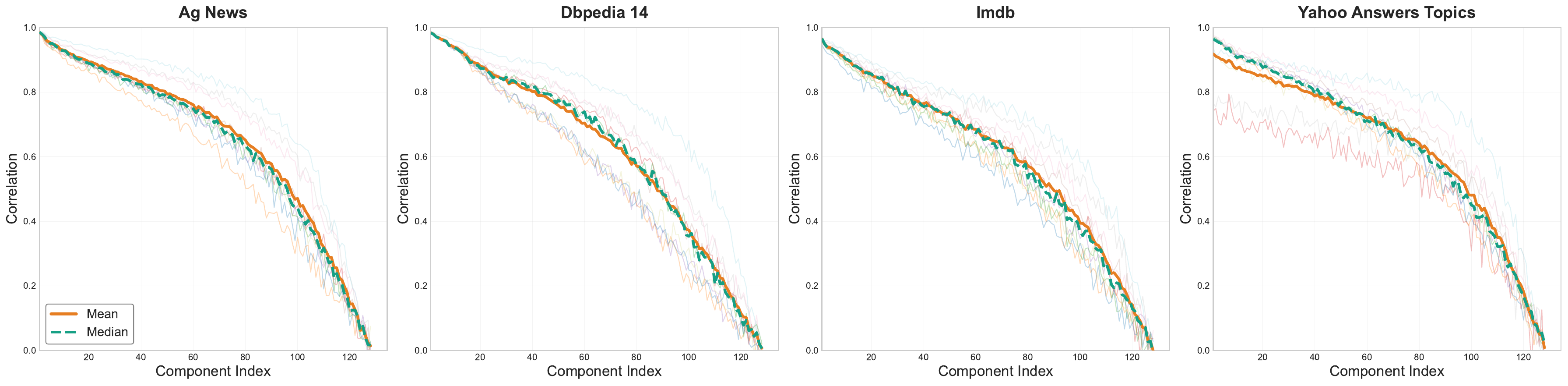} %
  \caption{SVCCA of five embedding model combinations at 128 components (test set)}
  \label{fig:svcca2}
\end{figure}

Figures~\ref{fig:svcca1} and~\ref{fig:svcca2} report SVCCA results for encoder-style embedding models evaluated across multiple datasets and model pairings, using projections onto the top 64 and 128 components, respectively. Across datasets and model combinations, we observe consistently high SVCCA correlations, indicating a strong shared low-rank subspace among embedding models.

\paragraph{Summary.}
Taken together, the CKA and SVCCA analyses provide evidence that independently trained embedding models learn compatible representations. Despite differences in training objectives, architectures, and fine-tuning procedures, these models preserve shared geometric structure that is amenable to linear alignment, motivating our subsequent investigation into whether such alignment suffices for downstream behavioral transfer.

\begin{figure}[h!]
  \centering
  \includegraphics[width=0.9\linewidth]{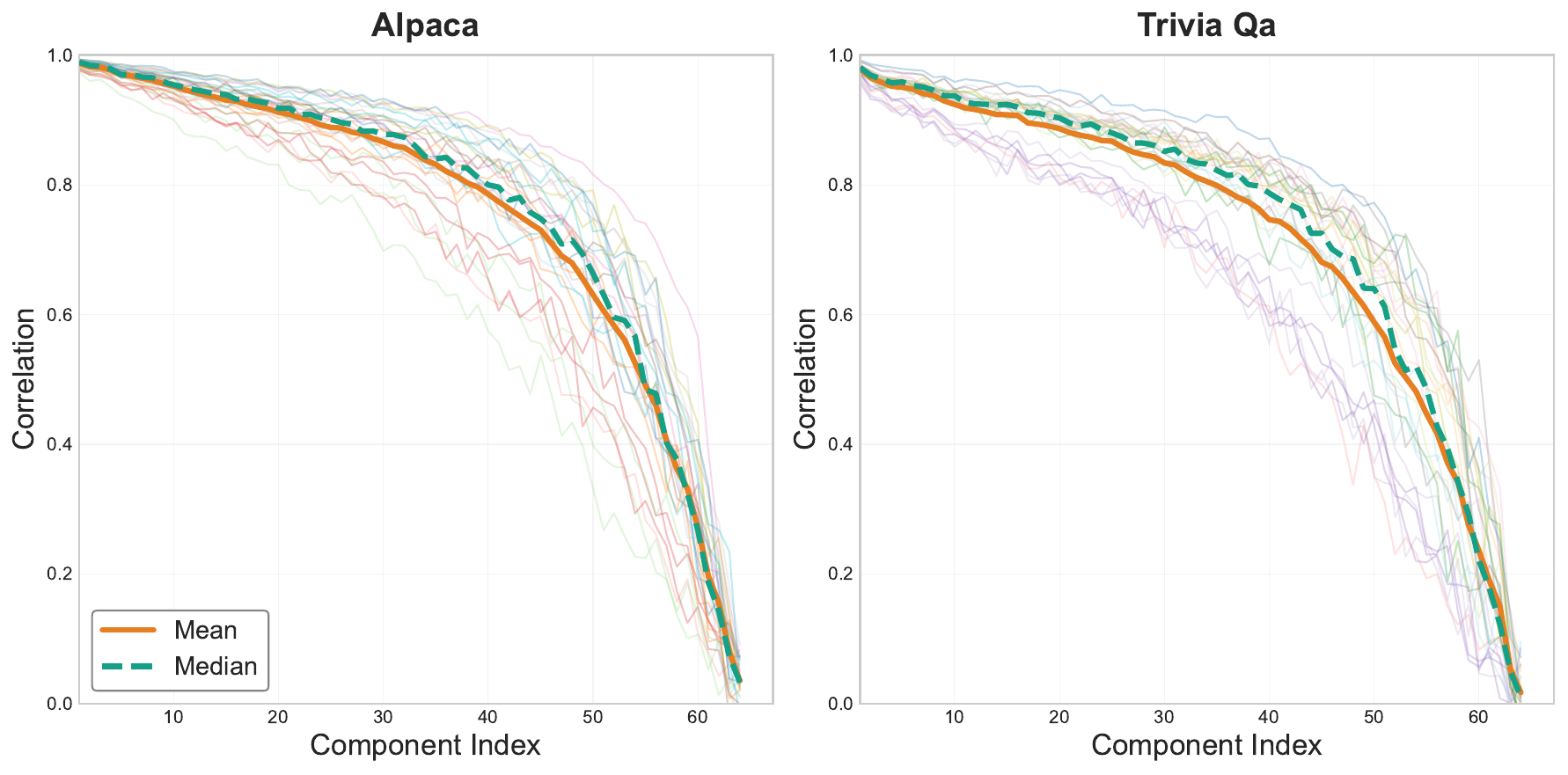} %
  \caption{SVCCA of eight instruction tuned \gls{llm} combinations at 64 components. Mean and median are bolded.  }
  \label{fig:svcca}
\end{figure}

\clearpage
\subsection{Task-Supervised Alignment Baseline }

To characterize the best-case performance of linear alignment, we evaluate an
\emph{in-distribution} setting where the mapping is trained directly on the task's
training split. We fit $(W^*,b^*)$ using paired embeddings from the training set
(computed by $g_A$ and $g_B$ on the same inputs) and evaluate on the held-out test
split using the fixed classifier $f_A(\cdot)$ trained on target embeddings. This
matches the alignment data to the downstream task distribution, providing an
upper-bound reference point for the distribution-mismatched results of
Section~\ref{experiments}, where the map is instead fit on out-of-distribution data.

\begin{table}[h]
\centering
\tiny
\renewcommand{\arraystretch}{1.0}
\setlength{\tabcolsep}{4pt}

\begin{tabular}{l l | c c | l c c}
\toprule
\textbf{Party A} & \textbf{Party B} &
\multicolumn{2}{c|}{\textbf{Classification Accuracy}} & \textbf{OOD} &
\multicolumn{2}{c}{\textbf{OOD AUROC}} \\
\textbf{(Target)} & \textbf{(Source)} &
\textbf{Baseline} &  Linear Map&
\textbf{Dataset} & \textbf{Baseline} & Linear Map \\
\midrule
\rowcolor{gray!10}
\multicolumn{7}{c}{\textbf{AG News (25\%)}} \\
\midrule
Gemini  & OpenAI  & 92.6 & 91.6 & MNLI & 0.908 & 0.891 \\
OpenAI  & Qwen    & 92.6 & 91.9 & MNLI & 0.953 & 0.913 \\
Cohere  & Gemini  & 91.9 & 91.5 & MNLI   & 0.875 & 0.885 \\
OpenAI  & Cohere  & 92.1 & 92.0 & MNLI   & 0.881 & 0.880 \\
Mistral & Qwen    & 92.6 &92.4  & MNLI   & 0.953 & 0.940 \\

\midrule
\rowcolor{gray!10}
\multicolumn{7}{c}{\textbf{DBpedia (7\%)}} \\
\midrule
Gemini  & OpenAI  & 98.9 & 98.6 & Yahoo  & 0.971 & 0.975 \\
OpenAI  & Qwen    & 99.0 & 98.7 & Yahoo  & 0.979 & 0.974 \\
Cohere  & Gemini  & 98.9 & 98.7 & Yahoo    & 0.973 & 0.980 \\
OpenAI  & Cohere  & 98.9 &98.7  & Yahoo    & 0.971 & 0.973 \\
Mistral & Qwen    & 99.0 & 98.9 & Yahoo    & 0.979 & 0.985 \\

\midrule
\rowcolor{gray!10}
\multicolumn{7}{c}{\textbf{Yahoo Answers Topics (10\%)}} \\
\midrule
Gemini  & OpenAI  & 70.9 &67.6  & SST-2 & 0.417 & 0.612 \\
OpenAI  & Qwen    & 71.5 & 68.6 & SST-2 & 0.623 & 0.491 \\
Cohere  & Gemini  & 70.2 &67.3  & SST-2   & 0.566 & 0.593 \\
OpenAI  & Cohere  & 69.7 & 69.4 & SST-2   & 0.481 & 0.444 \\
Mistral & Qwen    & 71.5 & 70.5 & SST-2   & 0.623 & 0.482 \\

\midrule
\rowcolor{gray!10}
\multicolumn{7}{c}{\textbf{MNLI (33\%)}} \\
\midrule
Gemini  & OpenAI  & 62.4 & 61.3 & IMDB & 0.367 & 0.092 \\
OpenAI  & Qwen    & 87.9 & 61.3 & IMDB & 0.905 & 0.693 \\
Cohere  & Gemini  & 65.0 & 56.8 & IMDB   & 0.111 & 0.263 \\
OpenAI  & Cohere  & 59.1 & 58.3 & IMDB   & 0.384 & 0.285 \\
Mistral & Qwen    & 87.9 & 78.5 & IMDB   & 0.905 & 0.650 \\

\midrule
\rowcolor{gray!10}
\multicolumn{7}{c}{\textbf{SST-2 (50\%)}} \\
\midrule
Gemini  & OpenAI  & 94.5 & 93.1 & AGNews & 0.826 & 0.801 \\
OpenAI  & Qwen    & 94.5 & 93.8 & AGNews & 0.859 & 0.848 \\
Cohere  & Gemini  & 94.4 & 91.7 & AGNews   & 0.819 & 0.870 \\
OpenAI  & Cohere  & 93.0 & 92.4 & AGNews   & 0.875 & 0.843 \\
Mistral & Qwen    & 94.5 & 93.7 & AGNews   & 0.859 & 0.826 \\

\midrule
\rowcolor{gray!10}
\multicolumn{7}{c}{\textbf{IMDB (50\%)}} \\
\midrule
Gemini  & OpenAI  & 94.9 & 95.9 & Amazon & 0.523 & 0.543 \\
OpenAI  & Qwen    &  95.7& 94.8 & Amazon & 0.535 & 0.535 \\
Cohere  & Gemini  &96.4  & 94.5 & Amazon   & 0.572 & 0.501 \\
OpenAI  & Cohere  & 94.8 & 94.9 & Amazon   & 0.379 & 0.534 \\
Mistral & Qwen    & 95.7 & 95.1 & Amazon   & 0.535 & 0.617 \\

\midrule
\rowcolor{gray!10}
\multicolumn{7}{c}{\textbf{TREC (16.7\%)}} \\
\midrule
Gemini  & OpenAI  & 95.4 & 94.0 & AGNews & 0.412 & 0.718 \\
OpenAI  & Qwen    & 96.0 & 95.6 & AGNews & 0.738 & 0.916 \\
Cohere  & Gemini  & 97.0 & 91.2 & AGNews   & 0.921 & 0.766 \\
OpenAI  & Cohere  & 96.4 & 94.4 & AGNews   & 0.954 & 0.802 \\
Mistral & Qwen    & 97.0 & 96.6 & AGNews   & 0.921 & 0.940 \\

\bottomrule
\end{tabular}

\label{tab:ood_results}
\end{table}

\section{Text Generation Evaluation Methodology}
To obtain a robust view of generation quality, we combine automated and human evaluation signals. We first use a dual LLM-as-a-judge protocol to score relevance, quality, and helpfulness at scale, then validate these trends with a blinded human study on a representative subset of generations. Finally, we analyze embedding-space similarity to characterize how mapped representations relate to both the source and target model feature spaces during generation.
\subsection{LLM-as-a-Judge Evaluation}

We employ a dual-judge framework using GPT-5.2 (\texttt{gpt-5.2}) and Claude 4 Sonnet (\texttt{claude-sonnet-4}). Both judges independently evaluate each response, and final scores are averaged across judges.

\paragraph{Evaluation Prompt}

\begin{verbatim}
You are an expert evaluator of AI-generated responses. Evaluate the quality 
of the following response to the given prompt.

Prompt: {prompt}
Response to evaluate: {response}

Please evaluate the response on a scale of 1-10 based on:
1. Relevance: Does the response address the prompt appropriately?
2. Quality: Is the response well-written, coherent, and accurate?
3. Helpfulness: Is the response useful and informative?

Provide your evaluation in JSON format:
{"score": <1-10>, "reasoning": "<explanation>", 
 "relevance": <1-10>, "quality": <1-10>, "helpfulness": <1-10>}
\end{verbatim}

The aggregate score is the mean of the three dimensions. Both judges operate at temperature 0.0 with JSON-formatted outputs.

\subsection{Human Evaluation}

To validate the fidelity of automated evaluators used throughout this section (LLM-as-a-judge scores and embedding-based similarity analyses), we additionally collect an independent human assessment of a subset of generations. We sample 200 prompt--response pairs, export them to a Google spreadsheet, and hide the source configuration (single-model vs.\ cross-model, as well as model identity) from the evaluator. Each example is rated on a 1--5 Likert scale across four criteria: proper English, quality, and coherence.

\paragraph{Sample Selection}

Samples are distributed across eight configurations to cover diverse performance levels.  We select 2 models from each of the LLM-as-a-judge categories (cross-model (poor), cross-model (good), single model (small, poor), single model (large, good))

\paragraph{Cross-Model (100 samples, 25 per pair):}
\begin{itemize}
    \item \textbf{Poor} (LLM-judge score $< 4.0$): Gemma-270M $\rightarrow$ Llama-3-8B, Qwen2.5-0.5B $\rightarrow$ Gemma-2-2B
    \item \textbf{Strong} (LLM-judge score $> 7.0$): Llama-3-8B $\rightarrow$ Qwen2.5-14B, Qwen2.5-7B $\rightarrow$ Llama-3-8B
\end{itemize}

\paragraph{Single-Model (100 samples):}
\begin{itemize}
    \item \textbf{Small} (50 samples): Qwen2.5-0.5B, Gemma-270M (25 each)
    \item \textbf{Large} (50 samples): Qwen2.5-7B, Llama-3-8B (25 each)
\end{itemize}

All examples are presented in randomized order and evaluated blind to model configuration. The spreadsheet contains columns for prompt, completion, and the ranking column.

\subsection{Embedding Space Analysis}

\begin{figure}
    \centering
    \includegraphics[width=0.55\textwidth]{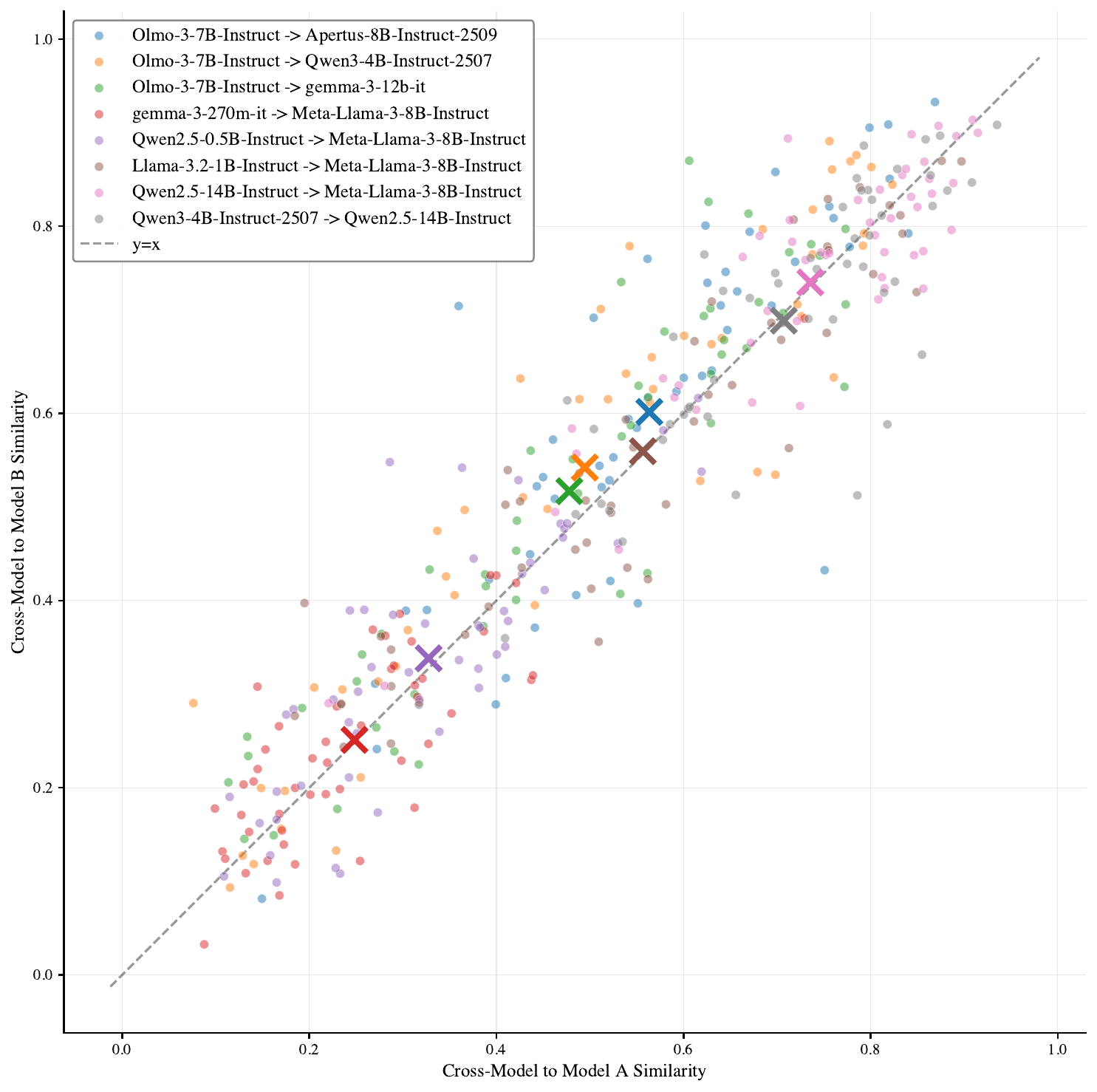}
    \caption{Embedding similarities between greedy-mapped text generation representations and cross-model text generation representations. }
    \label{fig:greedy_similarities}
\end{figure}

We analyze the cosine similarity between mapped representations and both source (Model A) and target (Model B) embeddings. Figure~\ref{fig:greedy_similarities} plots these similarities across eight model pairs, with each point representing a token position and X markers indicating mean values.

Model pairs exhibit substantial variation: larger models (Qwen2.5-14B $\rightarrow$ Llama-3-8B, Qwen3-4B $\rightarrow$ Qwen2.5-14B) achieve high similarity to both source and target (upper-right quadrant), while smaller source models (gemma-3-270m, Qwen2.5-0.5B) show lower correlation to both (lower-left). Most pairs cluster above the diagonal with mean similarities of 0.5-0.7 to both models, suggesting mapped representations form an intermediate space that partially retains source structure while incorporating target characteristics.

\subsection{Training Data Size Selection}

\begin{figure}[h]
    \centering
    \includegraphics[width=0.6\textwidth]{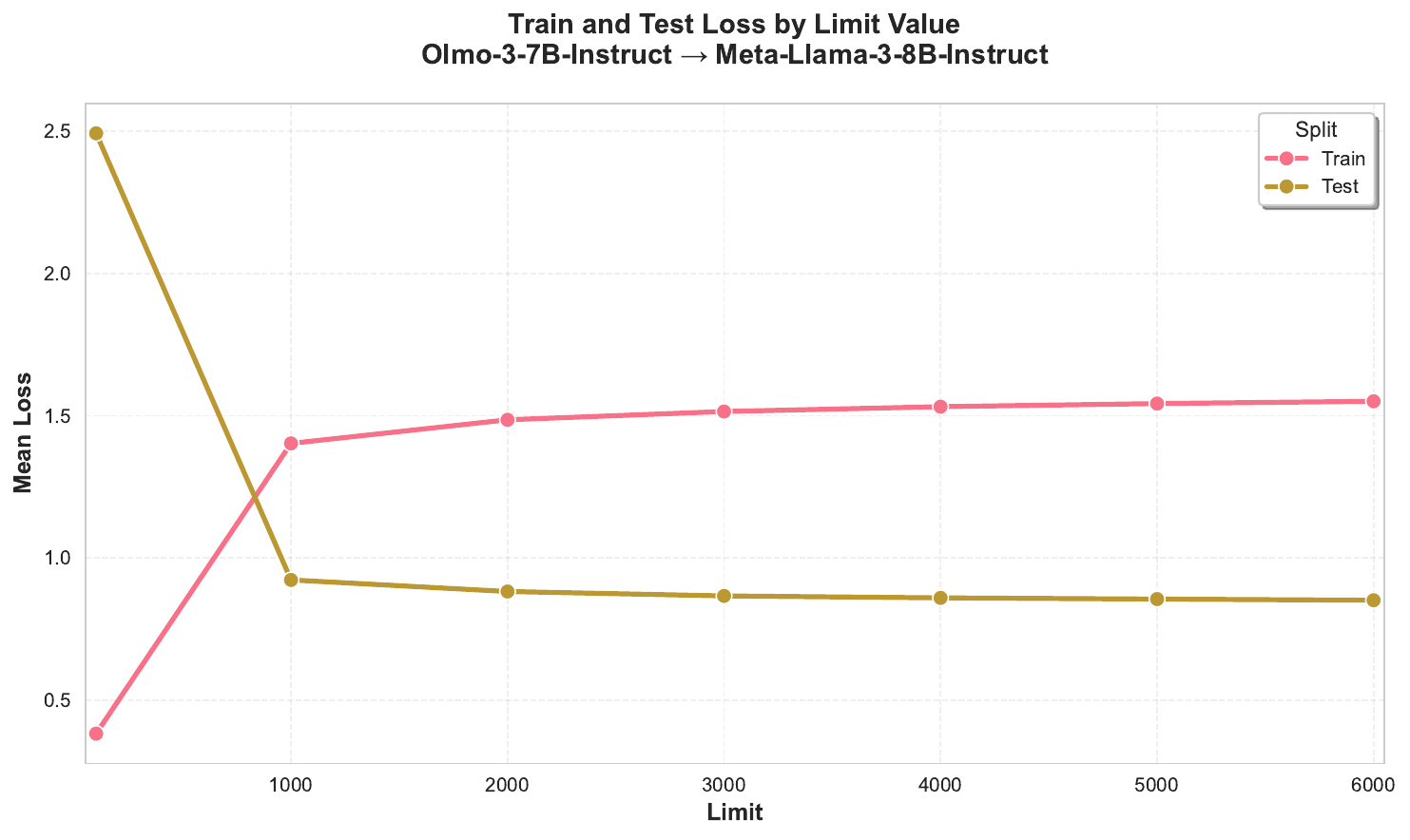}
    \caption{Training and test loss for cross-model alignment as a function of training dataset size. Test loss plateaus around 4,000 samples, while training loss continues to increase, suggesting diminishing returns beyond this point.}
    \label{fig:loss_by_limit}
\end{figure}

To determine an appropriate training dataset size for learning cross-model alignments, we conducted experiments varying the number of activation pairs used during mapper training. Figure~\ref{fig:loss_by_limit} shows the training and test loss curves for the Olmo-3-7B-Instruct $\rightarrow$ Meta-Llama-3-8B-Instruct alignment as a function of dataset size (ranging from 100 to 6,000 samples).

We observe that test loss (gold line) decreases rapidly from 100 to 1,000 samples, then plateaus around 0.86-0.87 for larger dataset sizes. In contrast, training loss (pink line) continues to increase monotonically, rising from approximately 1.41 at 1,000 samples to 1.54 at 6,000 samples. This divergence indicates overfitting to the training set with larger datasets.

Based on these results, we select \textbf{4,000 samples} as the standard training size for experiments, balancing computational efficiency with alignment effectiveness. This choice captures most of the performance gains while avoiding unnecessary computation and overfitting observed at larger dataset sizes.

\subsection{Token-Level Alignment}

Given a shared text corpus, we extract final-layer hidden states from both models and align them at the token level using character offsets. For each token in the source model (A) at position $i$ with character end offset $e_A^i$, we match it to the target model (B) token whose end offset equals or immediately follows:

$$j = \arg\min_{k} \{e_B^k \mid e_B^k \geq e_A^i\}$$

This produces aligned pairs $(h_A^i, h_B^j)$ used for training the affine transformation. The character-offset alignment accommodates tokenizer differences between models without requiring identical vocabularies.

\clearpage
\clearpage
\section{Text Generation Experimental Results}\label{text_gen_exp}

Table~\ref{tab:tokenizer_compatibility} presents comprehensive tokenizer compatibility metrics and LLM-as-a-Judge generation quality scores for all evaluated model pairs. We organize results into three categories: high-quality pairs (both models $\geq$2B parameters, score $\geq$3.5), low-quality pairs (both models $\geq$2B, score $<$2.0), and pairs involving small models ($<$2B parameters). For comparison, we include single-model baseline scores at the bottom of the table.

\textbf{High-quality pairs.} The top section shows 11 model pairs that achieve functional cross-model generation (LLM-judge scores 3.9--4.7). These pairs exhibit strong tokenizer compatibility: vocabulary overlap (Jaccard) $\geq$0.64 and exact token match rates $\geq$0.67. Same-family pairs (e.g., Qwen-7B $\rightarrow$ Qwen-14B, Mistral-Nemo $\rightarrow$ Apertus-8B) achieve near-perfect compatibility (Jaccard $\geq$0.999) and the highest generation quality. Cross-family pairs between Qwen and Llama models also perform well (Jaccard = 0.643, exact match = 0.925), demonstrating that models with shared tokenizer vocabularies enable effective linear alignment even across different model families.

\textbf{Low-quality pairs.} The middle section shows 7 pairs with both models $\geq$2B that produce poor-quality text (scores 1.1--1.9). These failures correlate strongly with low tokenizer compatibility: Jaccard $\leq$0.32 and exact match $\leq$0.67. Notably, all pairs involving Gemma models with non-Gemma models fail (Jaccard 0.057--0.063, exact match 0.227--0.238). The Mistral-7B model (vocabulary size 32K) also shows poor compatibility with models using larger vocabularies (131K--151K tokens).

\textbf{Small model pairs.} The bottom section demonstrates that small models ($<$2B) consistently fail at cross-model generation regardless of tokenizer compatibility. Even same-family pairs with perfect tokenization alignment (Llama-3.2-1B $\rightarrow$ Llama-3-8B: Jaccard = 1.0, exact match = 1.0) achieve only 1.83 LLM-judge score, suggesting that representational capacity is a prerequisite for successful linear alignment. The gemma-3-270M model fails universally (scores 1.0--1.06) across all target models, including same-family Gemma models with perfect tokenizer compatibility.

\textbf{Single-model baselines.} For context, we include native single-model performance on the same evaluation set. Cross-model generation quality (3.9--4.7 for high-quality pairs) falls between small models (2.8--4.1) and large models (6.5--7.1), representing a 60--70\% retention of baseline quality. 

\begin{table}
\centering
\small
\caption{Tokenizer Compatibility Metrics and Generation Quality for Model Pairs}
\label{tab:tokenizer_compatibility}
\begin{tabular}{@{}llccc@{}}
\toprule
\textbf{Model A} & \textbf{Model B} & \makecell{\textbf{Vocab}\\\textbf{Jaccard}} & \makecell{\textbf{Exact}\\\textbf{Match}} & \makecell{\textbf{LLM Judge}\\\textbf{Score}} \\
\midrule
\multicolumn{5}{l}{\textit{High-Quality Pairs (Score $\geq$ 3.5, both $\geq$2B)}} \\
\midrule
Qwen2.5-7B-Instruct & Qwen2.5-14B-Instruct & 1.000 & 1.000 & 4.68 \\
Qwen2.5-7B-Instruct & Meta-Llama-3-8B-Instruct & 0.643 & 0.925 & 4.47 \\
Qwen2.5-14B-Instruct & Meta-Llama-3-8B-Instruct & 0.643 & 0.925 & 4.43 \\
Mistral-Nemo-Instruct-2407 & Apertus-8B-Instruct-2509 & 0.999 & 1.000 & 4.35 \\
Meta-Llama-3-8B-Instruct & Qwen2.5-14B-Instruct & 0.643 & 0.925 & 4.31 \\
Mistral-Nemo-Instruct-2407 & Meta-Llama-3-8B-Instruct & 0.382 & 0.637 & 4.21 \\
Mistral-Nemo-Instruct-2407 & Qwen2.5-14B-Instruct & 0.316 & 0.672 & 4.13 \\
Qwen3-4B-Instruct-2507 & Qwen2.5-14B-Instruct & 1.000 & 1.000 & 4.07 \\
Qwen3-4B-Instruct-2507 & Meta-Llama-3-8B-Instruct & 0.643 & 0.925 & 3.99 \\
Qwen2.5-7B-Instruct & Apertus-8B-Instruct-2509 & 0.316 & 0.672 & 3.94 \\
Meta-Llama-3-8B-Instruct & Qwen2.5-7B-Instruct & 0.643 & 0.925 & 3.90 \\
\midrule
\multicolumn{5}{l}{\textit{Low-Quality Pairs (Score $<$ 2.0, both $\geq$2B)}} \\
\midrule
Mistral-Nemo-Instruct-2407 & Mistral-7B-Instruct-v0.2 & 0.067 & 0.151 & 1.89 \\
Qwen2.5-14B-Instruct & Mistral-Nemo-Instruct-2407 & 0.316 & 0.672 & 1.86 \\
Mistral-Nemo-Instruct-2407 & gemma-3-27b-it & 0.057 & 0.238 & 1.79 \\
Mistral-Nemo-Instruct-2407 & gemma-3-12b-it & 0.057 & 0.238 & 1.73 \\
Qwen3-4B-Instruct-2507 & gemma-3-12b-it & 0.063 & 0.227 & 1.68 \\
Qwen3-4B-Instruct-2507 & gemma-3-27b-it & 0.063 & 0.227 & 1.31 \\
Qwen2.5-7B-Instruct & Mistral-7B-Instruct-v0.2 & 0.061 & 0.151 & 1.13 \\
\midrule
\multicolumn{5}{l}{\textit{Pairs Involving Small Models ($<$2B)}} \\
\midrule
Llama-3.2-1B-Instruct & Meta-Llama-3-8B-Instruct & 1.000 & 1.000 & 1.83 \\
Llama-3.2-1B-Instruct & Qwen2.5-14B-Instruct & 0.643 & 0.925 & 1.53 \\
Qwen2.5-0.5B-Instruct & Meta-Llama-3-8B-Instruct & 0.643 & 0.925 & 1.18 \\
Llama-3.2-1B-Instruct & Qwen3-30B-A3B-Instruct-2507 & 0.643 & 0.925 & 1.17 \\
gemma-3-270m-it & gemma-3-27b-it & 1.000 & 1.000 & 1.06 \\
Llama-3.2-1B-Instruct & gemma-3-27b-it & 0.069 & 0.226 & 1.04 \\
gemma-3-270m-it & gemma-3-12b-it & 1.000 & 1.000 & 1.03 \\
gemma-3-270m-it & Meta-Llama-3-8B-Instruct & 0.069 & 0.226 & 1.03 \\
gemma-3-270m-it & Mistral-Nemo-Instruct-2407 & 0.057 & 0.238 & 1.02 \\
gemma-3-270m-it & Qwen3-30B-A3B-Instruct-2507 & 0.063 & 0.227 & 1.01 \\
gemma-3-270m-it & Qwen2.5-14B-Instruct & 0.063 & 0.227 & 1.00 \\
\midrule
\multicolumn{5}{l}{Single Models} \\
\midrule

DeepSeek-R1-Distill-Qwen-1.5B &--&--&--& 2.82 \\
google/gemma-3-270m-it &--&--&--&  3.40 \\
HuggingFaceTB/SmolLM2-1.7B &--&--&--&  3.68 \\
Qwen/Qwen2.5-0.5B-Instruct &--&--&--&  4.00 \\
google/gemma-2b &--&--&--&  4.08 \\
microsoft/Phi-3.5-mini-instruct (4B) &--&--&--&  5.36 \\
google/gemma-3-27b-it &--&--&--&  6.45 \\
Qwen/Qwen2.5-14B-Instruct &--&--&--&  6.65 \\
meta-llama/Meta-Llama-3-8B-Instruct &--&--&--&  7.05 \\
allenai/Olmo-3-7B-Instruct &-- &--&--& 6.70 \\
\bottomrule
\end{tabular}
\end{table}

\clearpage
\section{Cross-Model Text Generation: Sample Outputs}

\begin{table*}[h!]
\centering
\scriptsize
\renewcommand{\arraystretch}{1.2}
\begin{tabular}{|m{1\linewidth}|}
\hline
\multicolumn{1}{|c|}{\textbf{Llama-3-8B $\rightarrow$ Qwen2.5-14B}} \\
\hline
\textcolor{gray}{Input a list of strings and generate all possible permutations of the strings.
Input: a,b,c} \\
\ttfamily
The following are the possible permutations of the given list of strings:
1. "a", "b", "c" \\
2. "a", "c", "b" \\
3. "b", "a", "c" \\
4. "b", "c", "a" \\
5. "c", "a", "b" \\
6. "c", "b", "a" \\
\hline
\textcolor{gray}{Tell me one fascinating fact about dogs} \\
\ttfamily
One fascinating fact about dogs is that they have a unique, highly developed sense of smell. In fact, a dog's sense of smell is up to 1,000 to 10,000 more sensitive than a human's. This is due to the large number of specialized, highly specialized, and highly specialized, and highly specialized, and highly specialized, ...
 
\\
\hline

\textcolor{gray}{Provide a plan to improve the environment of your local park.} \\
\ttfamily
1. **Clean-up and Maintenance: \\
- Organize a community clean-up event to collect and dispose of any litter, debris, and hazardous materials.\\
- Establish a regular maintenance schedule for the park's landscaping, including regular watering, pruning, and planting of new plants and trees.\\
- Implement a recycling program for waste and encourage the use of reusable materials.\\
- Ensure proper maintenance of the park's infrastructure, including the maintenance of the park's pathways\\
\hline
\end{tabular}

\vspace{0.3cm}

\begin{tabular}{|m{1\linewidth}|}
\hline
\multicolumn{1}{|c|}{\textbf{Llama-3-8B $\rightarrow$ Qwen2.5-7B}} \\
\hline
\textcolor{gray}{Edit the following sentence: “To build the bridge we need construction equipment, including drills and jackhammers”} \\
\ttfamily
To build the bridge, we will need a variety of essential construction equipment, including, but not limited to, power tools such as drills and robust tools like the robust and powerful, 
\\
\hline
\textcolor{gray}{Evaluate the following expression: (6+2)*(2-2).} \\
\ttfamily
To evaluate the given expression, we need to follow the order of operations (PEM DAS) and perform the operations in the given order.

First, we need to evaluate the expression inside the parentheses: 2-2 = 0

Then, we need to add 6 and 2: 6 + 2 = 8

Now, we can use the result to perform the multiplication: 8 * 0 = 0

The final result of the given expression is 0. 
\\
\hline
\textcolor{gray}{Create a description of the characteristics of a capuchin monkey
} \\
\ttfamily
The capuchin monkey is a small, intelligent primate that belongs to the family Cebidae. It is native to the tropical forests of Central and South America, and is known for its agility, curiosity, and social behavior. Here are some of the key characteristics of the capuchin monkey:
\\
\hline
\textcolor{gray}{Research the differences between an object-oriented programming language and a procedural programming language.
} \\
\ttfamily
The primary difference between an object-oriented programming (OOP) and a procedural (or procedural-structured) programming (P) language is the way they structure and organize the code. 

In a procedural programming language, the program is organized as a series of procedures or functions that perform specific tasks. These functions are typically defined in a specific order, and the execution of the program follows a specific sequence. The main focus of a procedural language is on the process, and the code is organized to achieve a specific task. 

In contrast, a programming language that uses the principles of object-orientation (O) is based on the concept of
 
\\

\hline
\end{tabular}

\vspace{0.3cm}

\begin{tabular}{|m{1\linewidth}|}

\hline
\textcolor{gray}{Input a list of strings and generate all possible permutations: [``a'',``b'',``c'']} \\
\ttfamily
Here is a Python code that will generate all the possible (a, 1-1)
  
```
```
```
```
\\

\hline
\multicolumn{1}{|c|}{\textbf{Olmo-3-7B $\rightarrow$ Meta-Llama-3-8B-Instruct}} \\
\hline

\textcolor{gray}{
Create a list of reasons why having a pet can be beneficial.
} \\
\ttfamily
Here are some of the benefits of having a pet:
1. **Compani on in the Home: A pet can provide a loyal and loving companion, reducing the feeling of loneliness and loneliness. \\
2. Improved Mental Health: The presence of a pet has been shown to reduce stress, anxiety, and depression, and can improve overall mood and well- being. \\
3. Increased Physical Activity: C a t s and other pets can encourage their owners to exercise and be more active, which is beneficial for overall health. \\
4. Responsibility and Responsibility: C a re for a pet can help individuals develop a sense of responsibility\\
\\
\hline
\textcolor{gray}{
Create a 1-2 sentence summary of this research paper:
Input:
The paper, titled  "Exploring Voice-Based Human-Robot Interaction in Non-Technical Domains" describes the development of an interactive AI system that enables natural language-based human-robot interaction, with a focus on providing robotic assistance to homeowners.} \\
\ttfamily
The research developed a conversational AI system to enable users to interact with a home-based robot using natural language, allowing for easy and intuitive assistance in household tasks. The system is designed to facilitate human-robot interaction in a non-technical, everyday setting.
\\
\hline
\end{tabular}

\caption{Example prompts and corresponding responses generated using cross-model linear alignment for three model pairs. }
\label{tab:generation_examples}
\end{table*}

\clearpage
\section{Extended Related Works}

\textbf{Representational Similarity}
Complementary work on loss landscapes shows that overparameterized models often converge to functionally equivalent solutions up to symmetry transformations, despite large parameter-space variation \cite{entezari2021role, ainsworth2022git}. Methods for measuring representational similarity across neural networks include Canonical Correlation Analysis (CCA) \cite{raghu2017svcca}, centered kernel alignment (CKA) \cite{kornblith2019similarity}, and linear probing \cite{alain2018understandingintermediatelayersusing}. Recent work has explored the geometry of neural network representations \cite{huh2024platonic}, finding evidence of convergence toward shared representational structures across architectures and training procedures \cite{li2015convergent}. Studies on neural network stitching \cite{BansalNB21} demonstrate that layers from independently trained models can sometimes be connected via simple transformations, supporting the hypothesis that models learn aligned feature spaces.

\textbf{Transfer Learning} 
Transfer learning enables models pretrained on large corpora to adapt to new tasks with limited data through feature-based transfer \cite{pan2010survey} or fine-tuning \cite{devlin2018bert}. Knowledge distillation \cite{hinton2015distilling} trains compact student models to match teacher predictions, with variants including task-specific \cite{kim2016sequence}, self-distillation \cite{furlanello2018born}, and representation-level distillation \cite{romero2015fitnets,tian2019contrastive}.

\paragraph{Lossy Communication Between Models}
\citet{zou2026textpreservinglossytextcompression} compress text by strategically
deleting tokens and letting an LLM reconstruct the original from the retained
skeleton, keeping the channel textual for inspectability. We instead transmit
linearly aligned hidden representations, trading readability for compatibility
with encrypted computation.

\end{document}